%% file: main.tex
\documentclass[sigconf,arxiv]{acmart}
\renewcommand\footnotetextcopyrightpermission[1]{} % removes footnote with conference information in first column
\AtBeginDocument{%
  \providecommand\BibTeX{{%
    \normalfont B\kern-0.5em{\scshape i\kern-0.25em b}\kern-0.8em\TeX}}}

\usepackage{graphicx}
\usepackage{tikz}
 \usepackage{color}
\usepackage{colortbl} %use in the preamble
\usepackage{xcolor}
\usepackage{adjustbox}
\usepackage{subcaption}
\usepackage{tabularx}
 \usepackage{multirow}

\begin{document}

\title{Motif Learning in Knowledge Graphs\\ using Trajectories Of Differential Equations}

%\title{Neuro-Differential Embedding }
\author{Mojtaba Nayyeri}
\affiliation{%
 \institution{SDA Research University of Bonn, Germany  \\ Nature-Inspired Machine-Intelligence Research Group \\ InfAI Lab - Dresden, Germany}}
 \email{nayyeri@cs.uni-bonn.de}

\author{Chengjin Xu}
\affiliation{%
  \institution{SDA Research University of Bonn, Germany}}
\email{xu@@cs.uni-bonn.de}

\author{Jens Lehmann}
\affiliation{%
  \institution{SDA Research, University of Bonn, Germany \\ Fraunhofer IAIS, Dresden, Germany}
}
\email{jens.lehmann@iais.fraunhofer.de}

\author{Sahar Vahdati}
\affiliation{%
 \institution{Nature-Inspired Machine-Intelligence Research Group \\ InfAI Lab - Dresden, Germany } }
  \email{vahdati@infai.org}

% \affiliation{%

% \affiliation{%

% \affiliation{%

% \affiliation{%


\input{abstract}
% \begin{abstract}
%   A clear and well-documented \LaTeX\ document is presented as an
%   article formatted for publication by ACM in a conference proceedings
%   or journal publication. Based on the ``acmart'' document class, this
%   article presents and explains many of the common variations, as well
%   as many of the formatting elements an author may use in the
%   preparation of the documentation of their work.
% \end{abstract}




%\maketitle

\maketitle
\input{introduction}
\input{relatedwork}
\input{preliminaries}
\input{method}

\input{experiments}
\input{conclusion}

\bibliographystyle{ACM-Reference-Format}
\bibliography{main.bbl}

\appendix

\end{document}

%% file: abstract.tex
\begin{abstract}
    Knowledge Graph Embeddings (KGEs) have shown promising performance on link prediction tasks by mapping the entities and relations from a knowledge graph into a geometric space (usually a vector space).
    Ultimately, the plausibility of the predicted links is measured by using a scoring function over the learned embeddings (vectors). 
    Therefore, the capability in preserving graph characteristics including structural aspects and semantics, highly depends on the design of the KGE, as well as the inherited abilities from the underlying geometry.  
    Many KGEs use the flat geometry which renders them incapable of preserving complex structures and consequently causes wrong inferences by the models.
    To address this problem, we propose a neuro differential KGE that embeds nodes of a KG on the trajectories of Ordinary Differential Equations (ODEs).
    To this end, we represent each relation (edge) in a KG as a vector field on a smooth Riemannian manifold. 
    We specifically parameterize ODEs by a neural network to represent various complex shape manifolds and more importantly complex shape vector fields on the manifold. 
    Therefore, the underlying embedding space is capable of getting various geometric forms to encode complexity in subgraph structures with different motifs.
    Experiments on synthetic and benchmark dataset as well as social network KGs justify the ODE trajectories as a means to structure preservation and consequently avoiding wrong inferences over state-of-the-art KGE models. 
\end{abstract}

%Multi-relational graph embedding which aims at achieving effective representations with reduced low-dimensional parameters, has been widely used in knowledge base completion. 
%Although knowledge base data usually contains tree-like or cyclic structure, none of existing approaches can embed these data into a compatible space that in line with the structure.

 %including relations forming prevailing shapes such as path and loop simultaneously. 

%% file: introduction.tex
\section{Introduction}

%In the last decade, advances in theory and practice of 
Knowledge Graphs (KGs) have hugely impacted AI-based applications (on and off the Web) such as question answering, recommendation systems, and prediction services \cite{wang2017knowledge}.
A KG represents factual knowledge in triples of form (entity, relation, entity) e.g., (Plato, influences, Kant), in a large-scale multirelational graph where nodes correspond to entities, and typed links represent relationships between nodes.
%Nowadays, KGs stand as the backbone technology of many Web enterprises such as e-commerce, social networks, digital distribution, and online video-sharing platform. 
%Online social networks such as Twitter govern KGs for which this technology is being enormously influential by accurate, diverse, and explainable AI-based approaches in multiple applications such as question answering, recommendation, and perdition.
Although quantitatively KGs are often large-scale with millions of triples, this is nowhere near enough to capture the knowledge from real world.
%since capturing all of the knowledge is impossible.
To address this problem, various link prediction approaches have been used so far, among which link prediction using KG embedding (KGE) attracted a growing attention.
%A KGE model aims at measuring the likelihood of a link (ie.~a triple $(head,relation,tail)$) by mapping entities and relations of a KG from a symbolic domain to a geometric space (e.g.~a vector space).
KGEs map entities and relations of a KG from a symbolic domain to a geometric space (e.g.~a vector space).
KGEs employ a score function to perform the link prediction task which runs over the learned embedding vectors $(\boldsymbol{h,r,t})$ (bold represents a vector) of a triple $(h,r,t)$ and computes its plausibility (defining positive or negative triples).

\begin{figure*}
\centering % trim={<left> <lower> <right> <upper>}
\includegraphics[width=0.8\textwidth]{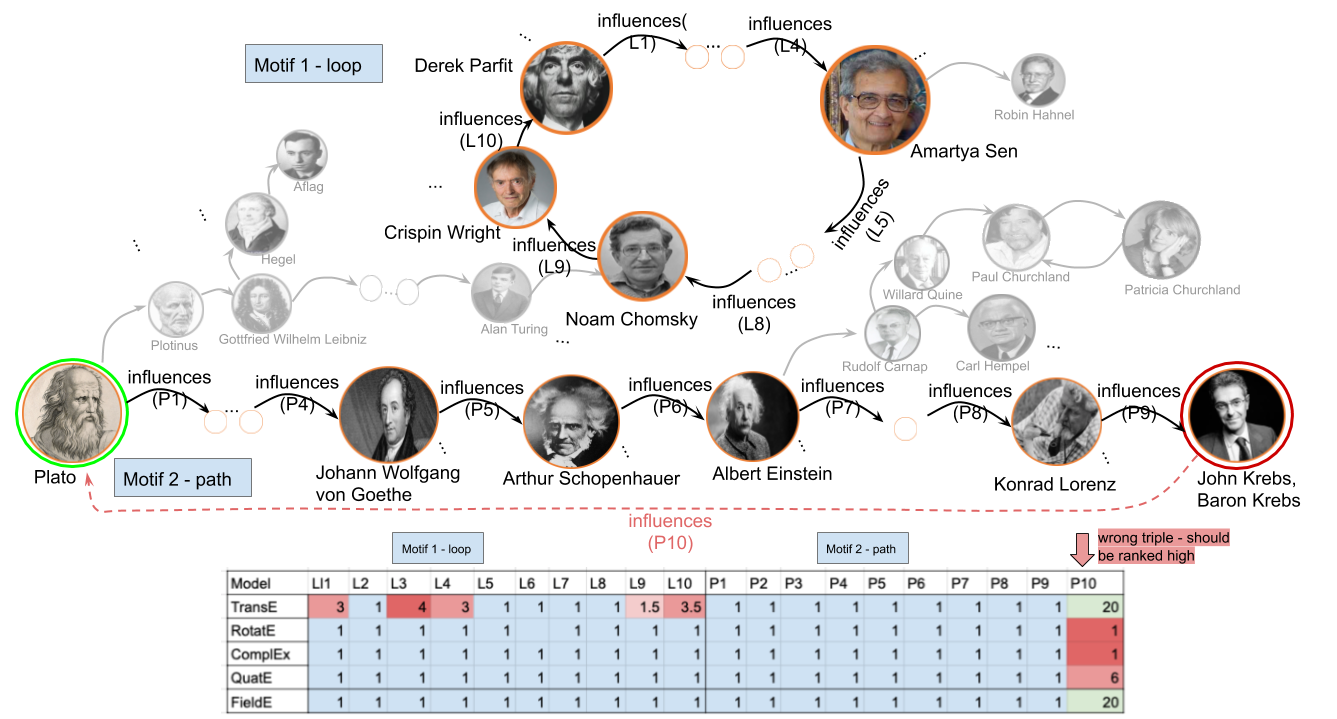}
  \caption{\textbf{Heterogeneous motifs by a single relation.} Illustration of path and loop generated by ``influences'' relation in YAGO.}
  \label{fig:path}
\end{figure*}

%    Therefore, the preservation potency of graph characteristics including structural aspects and semantics highly depends on the design of the KGE, as well as the inherited abilities from the underlying geometry.  
%Therefore high-level encoding characteristics such capability in capturing structural and semantical knowledge only remains in the triple level.

In this way, the major encoding capabilities of KGEs remain focused in the triple level.
Therefore, capturing collective knowledge both from the structural and semantical aspects for groups of triples (sungraphs) stays dependent on the inherited characteristics from the underlying geometries.
Within each geometry, the mathematical operations used in the score function make individual differences in encoding capability of KGEs. 
In most of the KGEs, while mapping each triple into a geometric space, naturally the subgraphs are also mapped.
%original:However, preserving subgraphs in a geometric space becomes challenging for KGEs, specially when the subgraph generate complex and heterogeneous structures and motifs. 
However, the extent to which the structure of subgraphs are preserved, remains limited to the characteristics of the considered geometric space, specially when a subgraph form complex and heterogeneous structures and motifs (i.e.~statistically significant shapes distributed in the graph). 
Major part of the state-of-the-art KGEs \cite{quate2019zhang,sun2019rotate, complex2016trouillon, bordes2013translating} are designed in flat geometries which do not intrinsically support structural preservation. 
Consequently, the returned scores of certain triples involved in such heterogeneous structures are measured inaccurately. 
Such heterogeneous structures are not only possible to occur between nodes connected with multiple relations but also can be caused by a single relation.
While many KGE models have been proposed, very few work have investigated on KGEs for discovery and encoding of heterogeneous subgraphs and motifs.
Here, we focus on the case of complex motifs in multirelational graphs that are caused by one \emph{single} relation e.g., ``follows'', or ``influences''. 
The nature of these relations leads to heterogeneous and completely different shapes in subgraphs, which is a prevailing case in many KGs.
%specially in KGs of social networks where lots of nodes are connected with a \emph{single} relation in different forms. 
In Figure \ref{fig:path}, we illustrate two possible subgraphs with different motifs generated by ``influences'' relation on the YAGO knowledge graph.
The first subgraph is captured as a loop structure which is constructed by 10 nodes. 
The second subgraph forms a path structure which is created by the same relation among another set of 10 nodes. 

Let us explore the following link prediction task by RotatE model on a toy KG with such motifs.
The assumption is that, the RotatE model returns wrong inferences when two subgraphs are structured in different shapes with a same relation. 
To show this let us represent the above motifs as $m_L$ and $m_P$ -- each with a set of 10 connected nodes ($e_1 \dots e_{10}$) with one relation ($r$). 
Therefore, (${e_1}^{m_L} \dots {e_{10}}^{m_L}$) represents the nodes of motif ($m_L$) where they form a loop, and (${e_1}^{m_P} \dots {e_{10}}^{m_{P}}$) correspond to the nodes of another subgraph with the same relation but forming a path. 
For each triple in this graph e.g. (${e_i}^{m_k}, r, {e_j}^{m_k}$), $k \in \{L,P\}$, the vector representation using RotatE is $\boldsymbol{e_i}^{m_k} \circ r  \approx \boldsymbol{e_j}^{m_k}$.
A complete representation of the motifs are then as following: 

\begin{equation*}
m_L:
\begin{cases}
({e_1}^{m_L}, r, {e_2}^{m_L}), \rightarrow \boldsymbol{e_1}^{m_L} \circ r  \approx \boldsymbol{e_2}^{m_L}, \\
({e_2}^{m_L}, r, {e_3}^{m_L}), \rightarrow \boldsymbol{e_2}^{m_L} \circ r \approx \boldsymbol{e_3}^{m_L},\\
\vdots \\
({e_9}^{m_L}, r, {e_{10}}^{m_L}), \rightarrow \boldsymbol{e_9}^{m_L} \circ r \approx \boldsymbol{e_{10}}^{m_L},\\
({e_{10}}^{m_L}, r, {e_1}^{m_L}). \rightarrow \boldsymbol{e_{10}}^{m_L} \circ r \approx \boldsymbol{e_1}^{m_L}.
% \\
% \mathclap{\rule{5cm}{0.4pt}}
% \\
% \midrule
\end{cases}  
% \begin{minipage}{.3\linewidth}
% \centering (1)~$\leftrightarrow$~(2)
% \end{minipage}%
\end{equation*}

\begin{equation*}
m_P:
\begin{cases}
({e_1}^{m_P}, r, {e_2}^{m_P}), \rightarrow \boldsymbol{e_1}^{m_P} \circ r  \approx \boldsymbol{e_2}^{m_P}, \\
({e_2}^{m_2}, r, {e_3}^{m_P}), \rightarrow \boldsymbol{e_2}^{m_P} \circ r \approx \boldsymbol{e_3}^{m_P}, \\
\vdots \\
({e_9}^{m_P}, r, {e_{10}}^{m_P}). \rightarrow \boldsymbol{e_9}^{m_P} \circ r \approx \boldsymbol{e_{10}}^{m_P}. 
% \\
% \mathclap{\rule{5cm}{0.4pt}}
% \midrule
\end{cases}
\end{equation*}

In order to compute the value of relation $r$, we start with the loop structure, by replacing the corresponding vectors of each triple in the following (e.g., by replacing the left side of $\boldsymbol{e_2}$ in the second triple equation of $m_L$, we get $\boldsymbol{e_1}^{m_L} \circ r \circ r = \boldsymbol{e_3}^{m_L}$). 
By doing this to the end, we conclude that $\boldsymbol{e_1}^{m_L} \circ r \dots \circ r = \boldsymbol{e_1}^{m_L}$ which means $ r \circ \dots \circ r = 1$ where $r$ is a complex number, therefore $\theta_{r} = \frac{2\pi}{10}$. 
This value is consistent for $r$ in the whole graph, therefore, this can be used to check whether the second motif is preserved. 
Here we replace the vectors as above and additionally include the value of $r$ driven from the first calculation.
After some derivations, we have $\boldsymbol{e_1}^{m_P} * e^\frac{20\pi i}{10} =  \boldsymbol{e_{10}}^{m_P} * r$.
With a simplification steps, this results in $\boldsymbol{e_1}^{m_P} = \boldsymbol{e_{10}}^{m_P} * r$, from which the model infers that the triple (${e_{10}}^{m_P},r, {e_1}^{m_P}$) is positive.
However, the actual shape of this motif is a path and this wrong inference shows a loop structure. 

%propagates to a wrong preservation of the original motif and changes the shape from an original path to a loop. 

The root cause of this problem lies in the entity-dependent nature of this relation.
However, most of the KGE models such as RotatE (as well as TransE, ComplEx, QuatE) consider relations independent of entities. 
This is visible in the heat map illustration in Figure \ref{fig:path}. 
As shown, $P_{10}$ is a wrong triple and should be ideally ranked as high as possible. 
All the models except TransE do not preserve the path structure and provide a low rank for this triple (infer it as a correct triple).
Although TransE preserves the path structure (ranking $P_{10}$ low), it fails in preserving the loop structure as it infers five triples to be wrong ($L_{1}$, $L_{3}$, $L_{4}$, $L_{9}$, and $L_{10}$). 
This is due to the limitation of their geometry. 
% This is due to the lack of ability in capturing the dependencies between different variables (e.g.~position of an object) and their higher order momentum (e.g.~ velocity, acceleration) which is directly enforced from the underlying geometry in which the model is designed. 
% While flat geometry cannot properly reflect complex graph structure, most of KGE models have been designed based on this geometry. 

%DEs in ML
In order to tackle this problem, our novel KGE models dubbed \emph{FieldE} employs DEs for embedding of KGs into a vector space. 
Differential Equations (DEs) are used as powerful tools with a general framework to accurately define connection between neighboring points laid on trajectories, implying the continuity of changes, and consequently describing the underlying geometry. 
%The broad and comprehensive characteristics of DEs in capturing the underlying geometry of data consumed by ML models have been well-studied recently in Machine Learning (ML) literature. 
This is specially important because the success of a KGE model depends on the way it correctly specifies the underlying geometry that describes the natural proximity of data samples in a geometric space \cite{mathieu2020riemannian}. 
Designing \emph{FieldE} with a well-specified geometry 
a) improves generalization, 
b) 
%reduces the overall number of parameters, and c)
increases the interpretability. 
First-order Ordinary Differential Equations (FO-ODEs) are special class of DEs which represents a vector field on a smooth Riemannian manifold.
%The most common use of differential equations in science is to model dynamical systems, i.e. systems that change in time according to some fixed rule. 
%For such a system, the independent variable is (for time) instead of, meaning that equations are written like.
Therefore, we particularly focus on FO-ODEs in designing our model.
\emph{FieldE} brings the power of DEs, embeddings and Neural Networks together and provides a fully comprehensive model capable of learning different motifs in subgraphs of a KG. 

%% file: relatedwork.tex
\section{Related Work}
Here, we collected prior work on the capability of embedding models in preserving subgraph structures, and semantically relational patterns.
We also discuss the related work about the motifs and the use of DEs for different encoding purposes. 
%DE and Geometry as well as their application in machine leaning with a focus on Neural DE, and natural science including physics, chemistry, control theory.

\textbf{Learning Relational Motifs.} 
The primary generation of embedding models includes a list of translation-based approaches where the encoding of motifs have only been discussed with the essence simple relational patterns such as 1-1 in TransE~\cite{bordes2013transe} and 1-many, many-1, and many-many in its follow up models~\cite{ji2015knowledge,lin2015learning,wang2014knowledge}.
RotatE~\cite{sun2019rotate} is the first KGE model where rotational transformations have been used for encoding of more complex patterns such as symmetry, transitive, composition, reflexive, and inversion which also create complex subgraphs. 
Another group of KGEs which are using element-wise multiplication of transformed head and tail namely DisMult~\cite{yang2014embeddingDistmult}, ComplEx~\cite{complex2016trouillon}, QuatE~\cite{zhang2019quaternion}, and RESCAL~\cite{nickel2011three}, also belong to rotation-based models and some use the angle of transformed head and tail for measuring the correctness of the predicted links. 
Apart from some partial discussions about the capability of the models in encoding of relational patterns, capturing the motifs have not been directly targeted in any of these models.
A recent KGE model named MuRP ~\cite{balazevic2019multi} sheds the lights into the capability of embedding models in learning hierarchical relations.
It proposes a geometrical approach for multi-relational graphs using Poincaré ball~\cite{ji2016knowledge}.
However, our work not only covers hierarchical relations but also focuses on different complex motifs (e.g. simultaneous path and loop) constructed by one relation in multi-relational graphs. 
In \cite{suzuki2018riemannian}, a very specific derivation of TransE model for encoding hierarchical relations have been proposed.
The existing embedding models have been compared to the four different versions of the proposed Reimanian TransE namely hyperbolic, spherical, and euclidean geometries.
Theses versions are designed for a very specific conditions in different geometries, while our model uses a neural network which generalizes cases beyond these.

\textbf{Learning with Differential Equations.}
%A different but close set of recent works have been focusing on directly learning structural representations either by KGEs or neural networks. 
In \cite{chen2018neural}, a family of deep neural network models has been proposed which parameterizes the derivative of a hidden state instead of the usual specification of a discrete sequence of hidden layers. 
%$ \mathbf{h}_{t+1} = \mathbf{h}_{t} + f(\mathbf{h}_{t},{\theta}_{t})$.
In this approach, ODEs are used in the design of the continuous-depth networks with the purpose of providing an efficient computation of the network output which brings memory efficiency, adaptive computation, parameter efficiency, scalable and invertible normalizing flows, and continuous time-series models. 
It is applied for supervised learning on an image dataset, and time-series prediction.
Few works use Lorenz model in their approaches \cite{bose2020latent,chamberlain2017neural} which are not about knowledge graphs in our context.. 
%Using ODEs, a vector field is defined which continuously transforms the state of neural network.
%$\frac{d\mathbf{h}(t)}{dt} = f((\mathbf{h}(t), t, {\theta})$.
%In this work, an end-to-end training of ODEs is considered for the experiments which highlights the following advantages of using DE:
%Using continuous normalizing flows, the density matching and maximum likelihood in training have been achieved. 
This work used ODEs in the proposed approach without considering knowledge graphs and embeddings for link prediction tasks.
%From the methodology point of view, time-seriesn is seen as a manifold, however in our approach we view entities as points on a continuous single relation on the underlying vector field. 
In another recent work, the continuous normalizing flows have been extended for learning Riemannian manifolds \cite{mathieu2020riemannian} in natural phenomena such as volcano eruptions.
In this work, the manifold flows are introduced as solutions for ODEs which makes the defined neural network independent of the mapping requirements forced by the underlying euclidean space.

% Guideline for application of DEs:
% https://www.analyzemath.com/calculus/Differential_Equations/applications.html

% https://www.asc.ohio-state.edu/physics/ntg/6810/readings/hjorth-jensen_notes2013_08.pdf

% Application of DEs in machine learning:

% Computer vision:
% https://dl.acm.org/doi/pdf/10.1145/166117.166151?casa_token=ScsMdYjmDtkAAAAA:966Kx0CGyt1ziLbAfyZtOU7y4PWK0BIE5YiHUMc5xmA_HMsMP4jBWobFvCxIcBHeh_s2TWHeEEGpzA

% Neural DE:

% https://papers.nips.cc/paper/7892-neural-ordinary-differential-equations.pdf

% Nickel work: continuous normalization flow

% review of embedding models based on geometry:
% Euclidean (transe, quate, complex, rotate, tucker)  and non-euclidean:
% Riemannian TransE: 
% https://openreview.net/forum?id=r1xRW3A9YX
% MURP, 
%https://arxiv.org/pdf/1905.01669v2.pdf
%Social Network Dataset

%% file: preliminaries.tex
\section{Preliminaries and Background}
This section provides the preliminaries of the Riemannian Geometry and its corresponding key elements as the required background for our model. 
The aim of this paper is to embed nodes (entities) of a KG on trajectories of vector fields (relation) laid on the surface of a smooth Riemannian manifold.
Therefore, we first provide the mathematical definitions \cite{peter2006} for \emph{manifold}, and \emph{Tangent Space} followed by introducing \emph{vector field} including differential equations. 

\paragraph{\textbf{Manifold}}
% Let $\mathcal{M}$ be a $d$-dimensional topological manifold. $\mathcal{M}$ is topological Hausdorff space with a countable base which is locally similar to linear space i.e.~locally homeomorphic to $\mathbb{R}^d.$ There are additionally the following components
A $d$-dimensional topological manifold denoted by $\mathcal{M}$ is a Hausdorff space with a countable base which is locally similar to a linear space.
More precisely, $\mathcal{M}$ is locally homeomorphic to $\mathbb{R}^d$ where:
%There are additionally the following components
\begin{itemize}
    \item For every point $p \in \mathcal{M}$, there is an open neighbourhood $U$ around $p$ and a homomorphism $\phi: U \xrightarrow[]{} V$ which maps $U$ to $V \subset \mathbb{R}^d.$ 
    $\phi$ is called chart or the coordinate system. 
    $\phi(p) \in \mathbb{R}^d$ is the image of $p \in U$, which is called the coordinates of $p$ in the chart.
    
    \item Let $\bigcup\limits_{i=1}^{n} U_i = \mathcal{M},\,\,\, i=1,\ldots,n,$ meaning that $\mathcal{M}$ is partitioned into $n$ parts denoted by $U_i, i=1,\ldots,n.$ 
    The set $\mathcal{P} = \{\phi_i | i=1,\ldots,n\},$ with domain $U_i$ for each $\phi_i$, is called atlas of $\mathcal{M}.$
\end{itemize}

\paragraph{\textbf{Tangent Space}}
If we assume a particle as a moving object on a manifold $\mathcal{M}$, then at each point $p \in \mathcal{M}$, the particle is free to move in various directions with velocity $v$. The set of all the possible directions that a particle goes by passing point $p$ form a space called the \emph{Tangent space}. 
%The formal definition of tangent space is given in the following.
Formally, given a point $p$ on a manifold $\mathcal{M}$, the tangent space $\mathcal{T}_p\mathcal{M}$ is the set of all the vectors which are tangent to all the curves, passing through point $p$. 
Let $\gamma: t \xrightarrow[]{} \mathcal{M}$ be a parametric curve on the manifold. 
$\gamma(t)$ maps $t \in [a,b]$ to $\mathcal{M}$ and passes through point $p$. 
A \emph{curve} in the local coordinate is $\phi\circ\gamma: t \xrightarrow[]{} \mathbb{R}^d$, from $t$ to manifold and then to the local coordinate $x = \phi\circ\gamma(t) = \phi(\gamma(t))$. 
%$x$ is the position on the local coordinate and the rate of change of position (velocity) on the point $p$ is computed by
$x$ represents the position of a particle on the local coordinate and the rate of the changes for different positions (velocity) on point $p$ is computed by
\begin{equation}
    v = \frac{d\phi\circ\gamma(t)}{dt}|_{t=t_0} = \big[\frac{dx^1}{dt}, \ldots, \frac{dx^d}{dt} \big]|_{t=t_0},
\end{equation}
where $p = \gamma(t_0), x^i(t)$ is i-th component of the curve on the local coordinates. 
The tangent vector $v$ is velocity at point $p$ in the local coordinate. If we specify every possible curve passing $p$, then all of the velocity vectors form a set called tangent space. 
In other words, the tangent space represents all possible directions in which one can tangentially pass through $p$. 
In order to move in a direction with the shortest path, exponential map is used.
The exponential map at point $p\in \mathcal{M}$ is denoted by $exp_p: \mathcal{T}_p \mathcal{M} \xrightarrow[]{} \mathcal{M}$. 
For a given small $\epsilon$ and $v \in \mathcal{T}_p \mathcal{M}$, the map $exp_p(\epsilon v)$ shows how a particle moves on $\mathcal{M}$ through the shortest path from $p$ with initial direction $v$ in $\mathbb{R}^d.$ 
A first order approximation of exponential map is given by $exp_p(v) = p + v$.

%Given the definition of manifold, curve, and tangent space, we can define the Riemannian manifold by introducing the metric tensor as follows.

\paragraph{\textbf{Riemannian Manifold}}
A tuple $(\mathcal{M}, g)$ represents a Reimanian manifold where $\mathcal{M}$ is a real smooth manifold with a \emph{Riemannian metric} $g$.
The function $g_p = g(p) = \langle.,.\rangle_p: \mathcal{T}_p \mathcal{M} \times \mathcal{T}_p \mathcal{M} \xrightarrow[]{} \mathbb{R}, p\in \mathcal{M}$  defines an inner product on the associated tangent space. 
The metric tensor is used to measure angle, length of curves, surface area and volume locally, and from which global quantities can be derived by integration of local contribution.

% \paragraph{Exponential Map}
% The exponential map at point $p\in \mathcal{M}$ is denoted by $exp_p: \mathcal{T}_p \mathcal{M} \xrightarrow[]{} \mathcal{M}$ and for small $t$, the map $exp_p(tv)$, where $v \in \mathcal{T}_p \mathcal{M}$, shows how to move on $\mathcal{M}$ as to take the shortest path from $p$ with initial direction $v$ in $\mathbb{R}^d.$ A first order approximation of exponential map is given by $exp_p(v) = p + v$.

\paragraph{\textbf{Vector Field}}
Let $x(t)$ be a temporal evolution of a trajectory on a $d$-dimensional smooth Riemannian manifold $\mathcal{M}$ and $\mathcal{T}\mathcal{M} = \bigcup\limits_{z\in \mathcal{M}}\mathcal{T}_z\mathcal{M}$ be a tangent bundle (the set of all tangent spaces on a Manifold). 
For a given Ordinary Differential Equation (ODE) 

\begin{equation}
    \frac{dx(t)}{dt} = f_{\theta}(x(t)),
\end{equation}

$f_{\theta}: \mathcal{M} \xrightarrow[]{} \mathcal{T}\mathcal{M}$ is a \textit{vector field} on the manifold, which shows the direction and the speed of movements along which the trajectory evolves on the manifold’s surface.

The vector field defines the underlying dynamics of a trajectory on a manifold and can get various shapes with different sparsity/density as well as various flows with different degrees of rotation. 
In field theory, this is formalized by two concepts of \textit{Divergence} and \textit{Curl}. 
\emph{Divergence} describes the density of the outgoing flow of a vector field from an infinitesimal volume around a given point $p$ on the manifold, while a \emph{Curl} represents the infinitesimal rotation of a vector field around the point. 
Here we present the formal definition of Divergence and Curl.
Without loss of generality, let $\boldsymbol{F} = f_{\theta_x} i + f_{\theta_y} j + f_{\theta_z} k $ be a continuously differentiable vector field in the Cartesian coordinates.
The \emph{divergence} of $\boldsymbol{F}$ is defined as follows

\begin{equation}
\begin{split}
    div \boldsymbol{F} = \bigtriangledown  .  \boldsymbol{F} 
    = (\frac{\partial}{\partial x}, \frac{\partial}{\partial y}, \frac{\partial}{\partial z}) . (f_{\theta_x}, f_{\theta_y}, f_{\theta_z}) 
    \\
    = \frac{\partial f_{\theta_x}}{\partial x} i + \frac{\partial f_{\theta_y}}{\partial y} j +  \frac{\partial f_{\theta_z}}{\partial z} k.
\end{split}
\end{equation}

At a given point $p\in \mathcal{M},$ if $div \boldsymbol{F}(p) > 0$, then the point is a \textit{source}, i.e.~the outflow on the point is more than the inflow. Conversely, if $div \boldsymbol{F}(p) < 0$, then the point $p$ is \textit{sink} i.e.~the inflow on the point is more than the outflow.
Given a continuous differentiable vector field $\boldsymbol{F} = f_{\theta_x} i + f_{\theta_y} j + f_{\theta_z} k$, a \emph{curl} of the vector field is computed as following

\begin{equation}
\begin{split}
&curl \boldsymbol{F} =  \bigtriangledown \times \boldsymbol{F} = \left| {\begin{array}{*{20}c}
i & j & k \\
\frac{\partial}{\partial x} & \frac{\partial}{\partial y} & \frac{\partial}{\partial z}\\ 
f_{\theta_x} & f_{\theta_y} & f_{\theta_z}\\ 
& \end{array} } \right| =\\ &(\frac{\partial f_{\theta_z}}{\partial y} - \frac{\partial f_{\theta_y}}{\partial z}) i + (\frac{\partial f_{\theta_x}}{\partial z} - \frac{\partial f_{\theta_z}}{\partial x}) j + (\frac{\partial f_{\theta_y}}{\partial x} - \frac{\partial f_{\theta_x}}{\partial y}) k.
\end{split}
\end{equation}

The curl is a vector with a limited length and a direction. 
The length of the vector shows the extent to which the vector field rotates and its direction specifies if the rotation is clock-wise or counterclockwise around the vector using right-hand rule.
%Note that, there is no convergence for the curl of a vector function. 

%% file: method.tex
\section{Method}
\begin{figure}
\centering % trim={<left> <lower> <right> <upper>}
\includegraphics[width=1.0\textwidth]{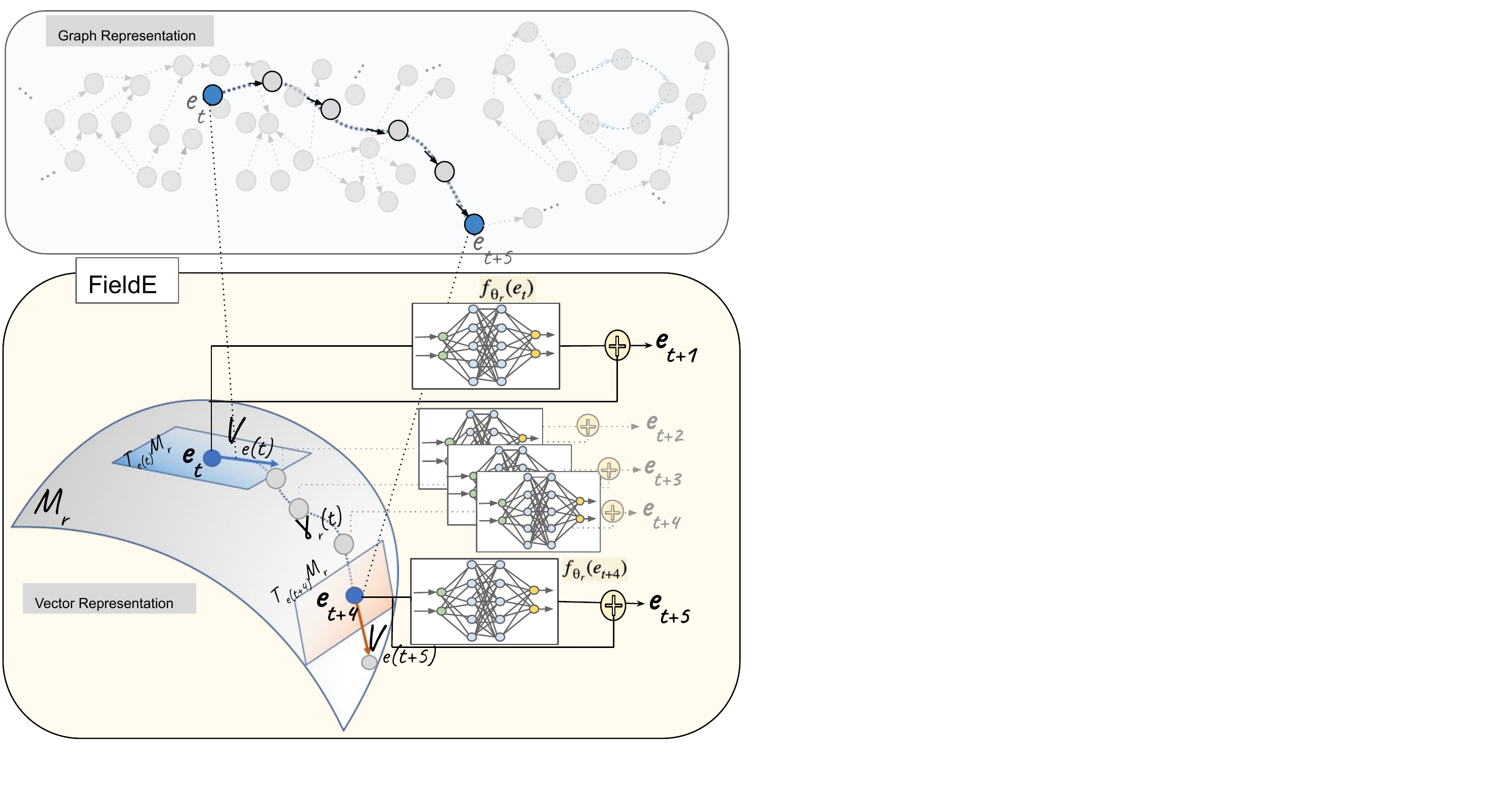}
  \caption{\textbf{The Architecture of FieldE Model}. The input to FieldE model is a knowledge graph which is shown in the upper part. A path structure is highlighted in the graph representation. The vector representation part (lower part) illustrates a trajectory of an ODE on a manifold. Nodes of the path are sequentially embedded on the trajectory guided by a neural network. 
  %FieldE provides a specific vector representation of the graph using ODEs further computes the plausibility of the triples using neural networks.
  }
  \label{fig:path}
\end{figure}

In this section, we propose \emph{FieldE}, a new KGE model based on Ordinary Differential Equations (ODEs). 
The formulation of \emph{FieldE} is presented in five folds:
\textit{relation formulation}, \textit{entity representation}, \textit{triple representation}, \textit{plausibility measurement}, and \textit{vector field parameterization} which are discussed in the remainder of this section. 

\paragraph{\textbf{Relation Formulation}}
\emph{\emph{FieldE}} represents each relation $r$ in a KG as a vector field ($f_{\theta_r}$) on a Riemannian manifold.
%$\theta_r$ denotes the parameters of the function $f_{\theta_r}$ which is constant by time.
$\theta_r$ denotes the parameters of the $f_{\theta_r}$ function and is constant by time.
If we assume, $\boldsymbol{x}(t)$ is a parametric trajectory that evolves by the changes of parameter $t \in \mathbb{R}^+$, the following ODE can be defined per each relation of the KG: 

\begin{equation}
    \frac{d\boldsymbol{x}(t)}{dt} = f_{\theta_r}(\boldsymbol{x}(t)), \,\,\, \boldsymbol{x}(t) \in \mathbb{R}^d,
    \label{relationODE}
\end{equation}

Given the above formulation, each relation of a KG forms a relation-specific geometric shape.
This is consistent with the nature of KGs where different relations form different motifs and patterns in the graph.

\paragraph{\textbf{Entity Representation}}
We represent each entity of a KG as a $d$-dimensional vector i.e.~$\boldsymbol{e} \in \mathbb{R}^d$.
The corresponding vectors are embedded on a trajectory of an ODE with an underlying relation-specific vector field. 
In other words, a sequence of entities (nodes) which are connected by a specific relation, are embedded sequentially through a trajectory laid on a relation-specific Riemannian surface. 
%This is actually consistent with the nature of multi-relational KGs where different relations might represent totally different structural patterns and motifs.

\paragraph{\textbf{Triple Learning}}
In order to formulate the steps for learning triples by \emph{FieldE}, let $e_t$ and $e_{t+1}$ be the two subsequent nodes (entities shown in the upper part of the Figure~\ref{fig:path}) of a graph connected by a relation $r$.
This actually forms as a triple $(e_t, r, e_{t+1})$ in a directed graph for each sequentially connected entities with relation $r$.
We consider $\boldsymbol{e}_t, \boldsymbol{e}_{t+1} \in \mathbb{R}^d$ as the embedding vectors for the subsequent entities $e_t$ and $e_{t+1}$ (shown in the lower part of the Figure~\ref{fig:path}).
Therefore, a triple in the vetor space can be modeled by time discretization
%Because the graph is directed and the two nodes (entities) are connected sequentially, we can model such a connection by providing a time discretization
(i.e.~$\frac{d\boldsymbol{x}(t)}{dt} \approx \frac{\Delta \boldsymbol{x}}{\Delta t} = \frac{\boldsymbol{x}(t+1) - \boldsymbol{x}(t)}{1} =  \boldsymbol{x}(t+1) - \boldsymbol{x}(t)$) over the equation \ref{relationODE} where $\boldsymbol{x}$ is replaced by entity embeddings ($\boldsymbol{e}$).
This gives the following equation
\begin{equation}
    \Delta \boldsymbol{e} = \boldsymbol{e}(t+1) - \boldsymbol{e}(t) = f_{\theta_r}(\boldsymbol{e}(t)),
    \label{relationODEDis}
\end{equation}

 where the time step is set to 1 i.e.~$\Delta t = 1.$ This is the first order approximation of the exponential map where $\boldsymbol{e}(t+1) = exp_{\boldsymbol{e}(t)}(f_{\theta_{r}}(\boldsymbol{e}(t))$. Therefore, Equation \ref{relationODEDis} shows how to move on the trajectory of a relation-specific manifold $\mathcal{M}_r$ to take the shortest path from the node $e(t)$ with initial direction $f_{\theta_{r}}(\boldsymbol{e}(t))$. 
 Note that the initial direction is dependent on the current node $e(t)$ and the relation $r$. 

For a given positive triple $(e_{t}, r, e_{t+1})$ in the graph, ($e(t)$ is shown by $e_t$ for simplicity from now onward), \emph{FieldE} learns the triple by optimizing the embedding vectors i.e.~$\boldsymbol{e}_t, \boldsymbol{e}_{t+1}$ as well as the relation-specific vector field function parameters i.e.~$\theta_r$ to fulfill
 
 \begin{equation}
     \boldsymbol{e}_{t+1} \approx \boldsymbol{e}_t + v_{e_t}, \,\,\, v_{e_t} = f_{\theta_r}(\boldsymbol{e}_t).
     \label{positiveapp}
 \end{equation}
 
 In order to regularize the current point $\boldsymbol{e}_t$ and the length of the step (velocity) $f_{\theta_r}(\boldsymbol{e}_t)$, we propose the following formulation instead of the Equation \ref{positiveapp}.
 Therefore, for a positive triple $(\boldsymbol{e}_{t},r,\boldsymbol{e}_{t+1}),$ \emph{FieldE} computes
 \begin{equation}
     \boldsymbol{e}_{t+1} \approx \eta \boldsymbol{e}_t \, + \, (1-\eta) v_{e_t},\,\,\, v_{e_t} = f_{\theta_r}(\boldsymbol{e}_t), \eta \in [0,1].
     \label{positiveappreg}
 \end{equation}

 Consequently, for a given negative triple $(\boldsymbol{e}'_{t},r,\boldsymbol{e}'_{t+1}),$ the following inequality should be fulfilled by \emph{FieldE}
 
 \begin{equation}
     \boldsymbol{e}'_{t+1} \neq \eta \boldsymbol{e}'_{t} + (1-\eta) v_{e_t}, \,\,\, v_{e_t} = f_{\theta_r}(\boldsymbol{e}'_t).
     \label{negativeappreg}
 \end{equation}

The Equations \ref{positiveappreg} and \ref{negativeappreg} are the optimal conditions to consider whether a triple is positive or negative. In the next part, we formulate the score function of \emph{FieldE} to measure the degree to which a triple is plausible.

\paragraph{\textbf{Plausibility Measurement}}
Given a triple $(e_t,r,e_{t+1})$, the plausibility of the triple is measured by
 
 \begin{equation}
     f_r(\boldsymbol{e}_t,\boldsymbol{e}_{t+1}) = -\|  \boldsymbol{e}_{t+1} -\eta  \boldsymbol{e}_t + (1-\eta) v_{e_t} \|,
     \label{ScoreDist}
 \end{equation}
 
 for distance-based version of our model, \emph{DFieldE}, and by
 
 \begin{equation}
     f_r(\boldsymbol{e}_t,\boldsymbol{e}_{t+1}) = \langle \boldsymbol{e}_{t+1} , \eta \boldsymbol{e}_t + (1-\eta) v_{e_t} \rangle,
     \label{ScoreSem}
 \end{equation}
 
for the semantic-matching version of our embedding model, \emph{SFieldE} where $v_{e_t} = f_{\theta_r}(\boldsymbol{e}_t)$. 
 
\paragraph{\textbf{Vector Field Parameterization}}
 
 The selection of the function $f_{\theta_r}$ is important as it determines the shape of the manifold as well as the shape of the underlying vector field. 
 In this paper, we propose two approaches for determining the vector field: a) we parameterize the vector field function $f_{\theta_r}$ by a neural network and propose a neuro-differential KGE models, b) we additionally propose linear version of our model where the vector field is modeled as a linear function. 
 Here we explain the two steps in detail:
\textbf{Neuro-FieldE}:
 Here we parameterize the vector field by a multi-layer feedforward neural network to approximate the underlying vector field 
 
 \begin{equation}
     f_{\theta_r}(\boldsymbol{e}_t) = \sum_{} w^o_i g(\sum_{} w_{ij}^L g(\sum_{} w_{jk}^{L-1} \ldots \sum_{} w_{pq}^2g(w_{qz}^1 \boldsymbol{e}_t + b_z^1)),
 \end{equation}
 where L is the number of hidden layer, $w^o$ denotes the output weight of the network and $w_{mn}^l$ is the weight connecting the $m$th node of the layer $l-1$ to the $n$th node of the $l$-th layer.

 Parametrizing the vector field with a neural network gives the model enough flexibility to learn various shape vector fields (representing complex geometry) from complex data. This is due to the advantage neural networks which are universal approximators (\cite{unihornik1989multilayer,unihornik1991approximation,uninayyeri2017universal}) i.e.~neural networks are capable of approximating any continuous function on a compact set. 
\textbf{Linear-FieldE}:
 Linear ODEs are a class of differential equations which have been widely used for several applications.
 Here we model the vector field as a linear function 
 
 \begin{equation}
     f_{\theta_r}(\boldsymbol{e}_t) = \mathcal{A}_r \boldsymbol{e}_t,
 \end{equation}
 
 where $\mathcal{A}_r$ is a $d \times d$ matrix. Depending the eigenvalues of $\mathcal{A}_r$, the vector field gets various shapes. 
 Below we present the theoretical analysis of \emph{FieldE} and its advantage over other state-of-the-art KGE models.
 
 \subsection{\textbf{Theoretical Analysis}}
 In this part, we theoretically analyse the advantages of the core formulation of \emph{FieldE} over other KGE models. 
 We first show that while other existing models such as RotatE, ComplEx face issues while learning on single-relational complex motifs (such as having path and loop with a single relation), \emph{FieldE} can easily model such complex structure. 
 Moreover, we show that our model subsumes several popular existing KGEs and consequently inherits their capabilities in learning various well-studied patterns such as symmetry, inversion, etc.
 
 \paragraph{Flexible Relation Embedding}
 Most of the existing state-of-the-art KGEs such as TransE, RotatE, QuatE, ComplEx etc., consider each relation of the KG as a constant vector to perform an algebraic operation such as translation or rotation.
 Therefore, the relation is entity-independent with regard to the applied algebraic operation. 
 For example, TransE considers a relation as a constant vector to perform translations as 
 
 \begin{equation}
     \boldsymbol{e_t + r}  \approx \boldsymbol{e_{t+1}}.
 \end{equation}
 Therefore, a relation-specific transformation (here translation) is performed in the same direction with the same length, regardless of the different  entities. 
 This causes an issue on the learning outcome of complex motifs and patterns. 
 To show this, let us consider a loop in a graph with a relation $r$ which connects three entities
 
\begin{equation}
\begin{split}
         &\boldsymbol{e_1 + r} \approx \boldsymbol{e}_2,\\
         & \boldsymbol{e_2 + r} \approx \boldsymbol{e_3},\\
         & \boldsymbol{e_3 + r} \approx \boldsymbol{e_1}.
\end{split}
\end{equation}
 
 After substituiting the first equation in the second one and comparing the result with the third equation, we conclude that $\boldsymbol{r} = \boldsymbol{0}$. 
 This is indeed problematic because embedding of all the entities will be the same i.e.~different entities are not distinguishable in the geometric space. 
 Now we prove that our model can encode loop without marginal issues. 
 
\begin{equation}
\begin{split}
         &\boldsymbol{e_1 + } f_{\theta_r}(e_1) \approx \boldsymbol{e}_2,\\
         & \boldsymbol{e_2 +} f_{\theta_r}(e_2) \approx \boldsymbol{e_3},\\
         & \boldsymbol{e_3 +} f_{\theta_r}(e_3) \approx \boldsymbol{e_1}.
         \label{ExampleEqs}
\end{split}
\end{equation}
 
In \emph{FieldE}, after substituiting the first equation in the second, and again substituting the result in the third equation, we obtain
 
 \begin{equation}
     f_{\theta_r}(\boldsymbol{e}_1) + f_{\theta_r}(\boldsymbol{e}_2) + f_{\theta_r}(\boldsymbol{e}_3) = \boldsymbol{0}.
     \label{ExampleEq}
 \end{equation}
 
 The above equation can be satisfied by \emph{FieldE} because neural networks with bounded continuous activation functions are universal approximators and universal classifiers \cite{unihornik1989multilayer, unihornik1991approximation, uninayyeri2017universal}. 
 Therefore, three points $\boldsymbol{e}_1, \boldsymbol{e}_2, \boldsymbol{e}_3$ can get the values by a well-specified neural network to hold the equality.
 
 We additionally show that our model can also embed a path structure with other three entities $e_4, e_5, e_6$ while preserving a loop structure with $e_1, e_2, e_3.$ 
 
\begin{equation}
\begin{split}
         &\boldsymbol{e_4 + } f_{\theta_r}(\boldsymbol{e}_4) \approx \boldsymbol{e}_5,\\
         & \boldsymbol{e_5 +} f_{\theta_r}(\boldsymbol{e}_5) \approx \boldsymbol{e_6},\\
         &\boldsymbol{e_6 +} f_{\theta_r}(\boldsymbol{e}_6) \neq \boldsymbol{e_4}.
         \label{ExampleNEs}
\end{split}
\end{equation}

 After substituting the first equation in the second equation, and again substituting the results in the third equation, we have 
 
 \begin{equation}
 \begin{split}
      f_{\theta_r}(\boldsymbol{e}_4) + f_{\theta_r}(\boldsymbol{e}_5) + f_{\theta_r}(\boldsymbol{e}_6) \neq \boldsymbol{0}. 
      \label{ExampleNEq}
\end{split}
 \end{equation}
 
 Because $\boldsymbol{e}_1, \ldots, \boldsymbol{e}_6$ are distinct points in the domain of the function fulfilling Equations \ref{ExampleEq}, and \ref{ExampleNEq}, there is a neural network that approximates these functions due to the universal approximation ability of the underlying network. 
 Therefore, \emph{FieldE} can learn two different sub-graph structures with the same relation. 
 
The state-of-the-art KGE models like TransE, RotatE, ComplEx and QuatE are not capable of learning the above structure because they always model the initial direction of relation-specific movement.
This is only to be dependent on the relation and ignore the role of entities in moving to the next node of the graph. 
Such limitation leads to wrong inferences when the graph contains complex motifs and patterns.
 
 \paragraph{Subsumption Of Existing Models}
 
Here, we prove that \emph{FieldE} subsumes popular KGE models and consequently inherits their learning power.
 
 \begin{definition}[from~\cite{kazemi2018simple}]
A model $M_1$ subsumes a model $M_2$ when any scoring over triples of a KG measured by model $M_2$ can also be obtained by model $M_1$.
\label{def:exre}
\end{definition}
 
\begin{proposition} \emph{DFieldE} subsumes TransE and RotatE. \emph{SFieldE} subsumes ComplEx and QuatE.
\end{proposition}

Because \emph{FieldE} subsumes existing models, it consequently inherits their advantages in learning various patterns including symmetry, and anti-symmetry, transitivity, inversion and composition. Moreover, because ComplEx is fully expressive and it is subsumed by Neuro-\emph{SFieldE}, 
we conclude that Neuro-\emph{SFieldE} is also fully expressive. 
Beside modeling common patterns, \emph{FieldE} is capable of learning more complex patterns and motifs such as having various motifs on a single relation e.g.~having loop and path with one relation.

\begin{proof}
Here we prove that \emph{DFieldE} subsumes TransE. 
The \emph{FieldE} assumption is
\begin{equation*}
    \boldsymbol{e}_{t+1} \approx \boldsymbol{e}_{t} + f_{\theta_r}(\boldsymbol{e}_t).
\end{equation*}
If we set $f_{\theta_r} = \boldsymbol{r}$ (constant vector field), then we have 
$\boldsymbol{e_{t+1}} \approx \boldsymbol{e}_t + \boldsymbol{r} $ which is the assumption of the TransE model for triple learning.
\end{proof}

\begin{proof}
We now prove that \emph{DFieldE} subsumes RotatE.
The RotatE assumption is

\begin{equation}
    \boldsymbol{e}_{t+1} \approx \boldsymbol{e}_{t} \circ \boldsymbol{r},
    \label{RotatEAssumption}
\end{equation}

where entities and relations are complex vectors and the modulus of each dimension of the relation vector is 1 i.e.~$|\boldsymbol{r}| = 1$. 
In the vector form, the equation \ref{RotatEAssumption} can be written in real (rotation) matrix-vector multiplication as following

\begin{equation*}
    \boldsymbol{e}_{t+1}^v \approx \boldsymbol{R}_r \boldsymbol{e}_{t}^v,
\end{equation*}

where $\boldsymbol{R}_r$ is a rotation matrix and $\boldsymbol{e}_{t}^v$ represents the vector representation of complex numbers (with two components of real and imaginary). 
Given the assumption of \emph{DFieldE} i.e.~
$\boldsymbol{e}^v_{t+1} \approx \eta \boldsymbol{e}^v_t + (1-\eta) f_{\theta_r}(\boldsymbol{e}^v_t),$ and setting $\eta = 0,$ and $ f_{\theta_r}(\boldsymbol{e}_t) = \boldsymbol{R}_r \boldsymbol{e}_{t}$, the assumption of RotatE is obtained. 
We conclude that, the RotatE model is a special case of \emph{DFieldE}.
\end{proof}

% \begin{table}[h!]
% \caption{Statistical information of the datasets. Number of entity and Number of relation present in the datasets are represented based on division between training, test triple, and validation sets.}
% %\begin{adjustbox}{width=\textwidth,center}
% \begin{tabular}{p{0.1\textwidth}|p{0.1\textwidth}|p{0.1\textwidth}|p{0.1\textwidth}|p{0.1\textwidth}|p{0.1\textwidth}}
% \hline
% Dataset & No.entity & No.relation & Training & Test & Validation  \\ \hline
% YAGO3-10 & 123k & 37 & 1m & 5k & 5k \\ \hline
% FB15k-237 & 15k & 237 & 272k & 20k & 18k \\ \hline
% \end{tabular}
% %\end{adjustbox}
% \label{table:table_dataset_stats}
% \end{table}

\begin{proof}
Here we present the proof of subsumption of ComplEx model.
The \emph{SFieldE} uses the following score function

\begin{equation*}
    f_r(\boldsymbol{e}^v_t,\boldsymbol{e}^v_{t+1}) = \langle \boldsymbol{e}^v_{t+1} , \eta \boldsymbol{e}^v_t + (1-\eta) f_{\theta_r}(\boldsymbol{e}^v_t)  \rangle.
\end{equation*}

After setting $\eta=0,$ we have

\begin{equation}
    f_r(\boldsymbol{e}^v_t,\boldsymbol{e}^v_{t+1}) = \langle \boldsymbol{e}^v_{t+1},  f_{\theta_r}(\boldsymbol{e}^v_t)  \rangle.
    \label{ReducedScore}
\end{equation}

Now let us focus on the score function of ComplEx which is

\begin{equation}
    f_r(\boldsymbol{e}_t,\boldsymbol{e}_{t+1}) = Re(\langle \bar{\boldsymbol{e}}_{t+1} ,  \boldsymbol{r},   \boldsymbol{e}_t\rangle).
    \label{ScoreComplEx}
\end{equation}

We represent the above equation in vectored version of complex numbers as following

\begin{equation}
    f_r(\boldsymbol{e}^v_t,\boldsymbol{e}^v_{t+1}) = \langle \boldsymbol{e}^v_{t+1} ,  \alpha_r \boldsymbol{R}_r   \boldsymbol{e}^v_t\rangle.
    \label{VecScoreComplEx}
\end{equation}

We can see if $f_{\theta_r}(\boldsymbol{e}_t) = \alpha_r \boldsymbol{R}_r \boldsymbol{e}^v_t$ in Equation \ref{ReducedScore}, we obtain the score of the ComplEx model in the vectorized in Equation \ref{VecScoreComplEx}. Therefore, ComplEx is also a special case of \emph{SFieldE}.

\end{proof}

\begin{proof}
Here we show that \emph{SFieldE} subsumes QuatE. QuatE uses the following formulae for the score function

\begin{equation}
    f_r(\boldsymbol{e}_t,\boldsymbol{e}_{t+1}) = \boldsymbol{e}_{t+1} .  \boldsymbol{r} \otimes  \boldsymbol{e}_t,
    \label{ScoreQuatE}
\end{equation}

where $\otimes, .$ show the Hamilton product and element-wise product between two quaternion vectors. Similarly to RotatE, the Equation \ref{ScoreQuatE} can be written in matrix vector multiplication as follows

\begin{equation}
    f_r(\boldsymbol{e}^v_t,\boldsymbol{e}^v_{t+1}) = \langle \boldsymbol{e}^v_{t+1},  \boldsymbol{R}_r \boldsymbol{e}^v_t \rangle, 
\end{equation}

where $\boldsymbol{R}_r$ is a $4d \times 4d$ matrix and $\boldsymbol{e}^v_t$ is a vectorized version of quaternion numbers.

Here, we show that the above equation can be constructed by the score function of \emph{SFieldE} which is

\begin{equation*}
    f_r(\boldsymbol{e}^v_t,\boldsymbol{e}^v_{t+1}) = \langle \boldsymbol{e}^v_{t+1} , \eta \boldsymbol{e}^v_t + (1-\eta) f_{\theta_r}(\boldsymbol{e}^v_t)  \rangle.
\end{equation*}

After setting $\eta=0,$ we have

\begin{equation}
    f_r(\boldsymbol{e}^v_t,\boldsymbol{e}^v_{t+1}) = \langle \boldsymbol{e}^v_{t+1} ,  f_{\theta_r}(\boldsymbol{e}^v_t)  \rangle,
\end{equation}

which will be same as the score function of QuatE in vectorized form if the vector Field is set to
$f_{\theta_r}(\boldsymbol{e}^v_t) = \boldsymbol{R}_r \boldsymbol{e}^v_t.$ 
Therefore, \emph{SFieldE} subsumes QuatE model, as well.
\end{proof}

%% file: experiments.tex
\section{EXPERIMENTS AND RESULTS}
%\begin{adjustbox}{width=\columnwidth,center}
In this section, we provide the results of evaluating FieldE 's performance in comparison to already existing state-of-the-art embedding models.
With a systematic analysis, we selected a list of KGEs to compare our model with, the list includes TransE, RotatE, TuckEr, ComplEx, QuatE, Dismult, ConvE, and MuRP.

\subsection{Experimental Setup}

\paragraph{\textbf{Evaluation Metrics}}
We consider the standard metrics for compassion in KGEs namely Mean Reciprocal Rank (MRR), and Hits@n ($n = 1, 3, 10$).
MRR is measured by $\sum_{j=1}^{n_t} \frac{1}{r_j}$, where $r_j$ is the rank of the $j$-th test triple and $n_t$ - the number of triples in the test set.
Hits@n is the number of testing triples which are ranked less than n, where n can be 1, 3, and 10. 

\paragraph{\textbf{Datasets}}
We run the experiments on two standard datasets namely FB15k-237 \cite{toutanova2015observed}, and YAGO3-10 \cite{mahdisoltani2013yago3}. 
Statistics of these datasets including the number of their entities and relations as well as the split of train, test, and validation sets are shown in Table \ref{tab:dssplit}.

\begin{table}[h!]
	\centering 
		\caption{\textbf{Dataset Statistics.} Number of entities and relations as well as the split of datasets.} 
	\begin{tabular}{@{}l |ccccc@{}}
		\toprule 
		\textbf{Dataset} & \textbf{\#Ent.} & \textbf{\#Rel.} &\textbf{\#Train} & \textbf{\#Valid.}  & \textbf{\#Test} \\\midrule
         YAGO3-10 & 123k & 37 & 1m & 5k & 5k \\ \hline 
         FB15k-237 & 15k & 237 & 272k & 20k & 18k \\ 
		\bottomrule
	\end{tabular}
	\label{tab:dssplit}
\end{table}

\paragraph{\textbf{Hyperparameter Search}}
We implemented our models in the Python using PyTorch library. We used Adam as the optimizer and tune the hyperparameters based on validation set. The learning rate ($r$) and batch size ($b$) are tuned on the set $r = \{0.0002, 0.002, 0.02, 0,1\}$, $b = \{512, 1024\}$ respectively. The embedding dimension $d$ is fixed to $100$ for YAGO3-10 and 1000 for FB15k-237.
We set the number of negative sample to $100$ for B15K-237 and $500$ for YAGO3-10, and used adversarial negative sampling for our model as well as the other models we have re-implemented. 
We presented two versions of FieldE namely DFieldE and SFieldE. DfieldE uses distance function to compute the score of a triple (see equation \ref{ScoreDist}). On the other SfieldE uses inner product for score computation \ref{ScoreSem}. Each of the above version of FieldE can either used Neural Network to approximate the vector field or use an explicit linear function as a vector field.  For the neural network based FieldE we add the prefix "N" to the beginning of the name of our model (either NDFieldE or NSFieldE). For linear version of FieldE, we use "L" as prefix in the name of the model (either LDFieldE or LSFieldE). For the Neural version of FieldE, we used a neural network with two hidden layers with $500, 100$ hidden nodes for YAGO3-10 and $100, 100$ for FB15K-237. We fixed the parameter $\eta$ to $1$ in equations $\ref{ScoreDist}$ and $\ref{ScoreSem}$. The details of the optimal hyperparameters are reported in the Table \ref{tab:hyper}.

\begin{table*}[h!]
	\centering 
		\caption{The optimal parameter setting for \emph{FieldE} is given for six different metrics.}
	\begin{tabular}{@{}l |ccccccc@{}}
		\toprule 
		\textbf{Dataset} & \textbf{dimension.} & \textbf{learning rate.} &\textbf{batch size.} & \textbf{hidden nodes.} & \textbf{active function}  & \textbf{neg.sample} \\\midrule
         YAGO3-10 & 100 & 0.002 & 512 & (500,100)&  tanh & 500 \\ \hline 
         FB15k-237 & 1000 & 0.1  & 1024 & (100,100) & tanh & 100 \\ 
		\bottomrule
	\end{tabular}
	\label{tab:hyper}
\end{table*}

\subsection{Results}
\begin{figure*}[h!]
	\centering 
	\begin{subfigure}[b]{0.31\textwidth}
		\includegraphics[width=\linewidth]{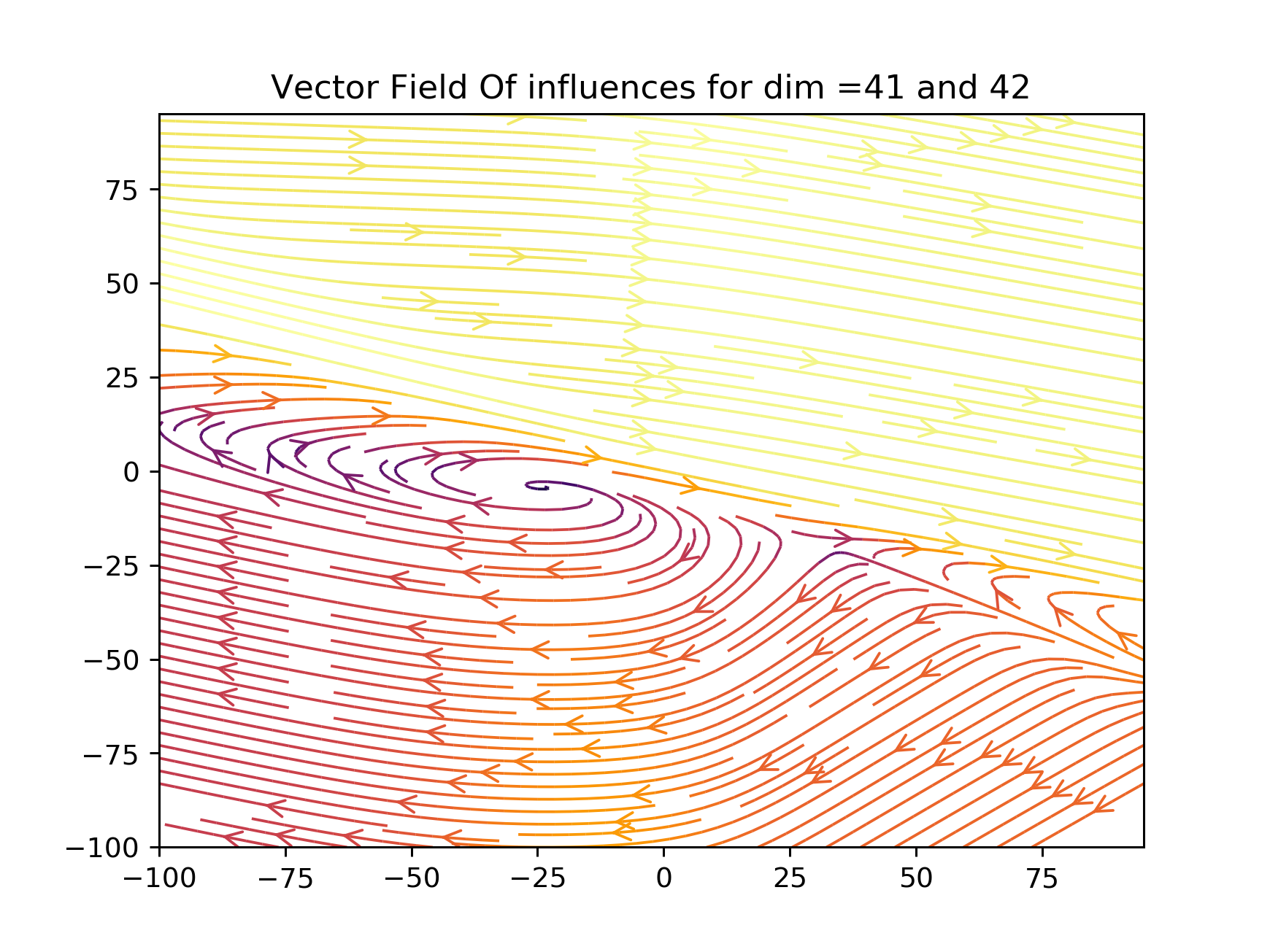}
	     \caption{{\scriptsize some people influencing many (loop and path).}}
		\label{fig:11}
	\end{subfigure}%
	\begin{subfigure}[b]{0.31\textwidth}
		\includegraphics[width=\linewidth]{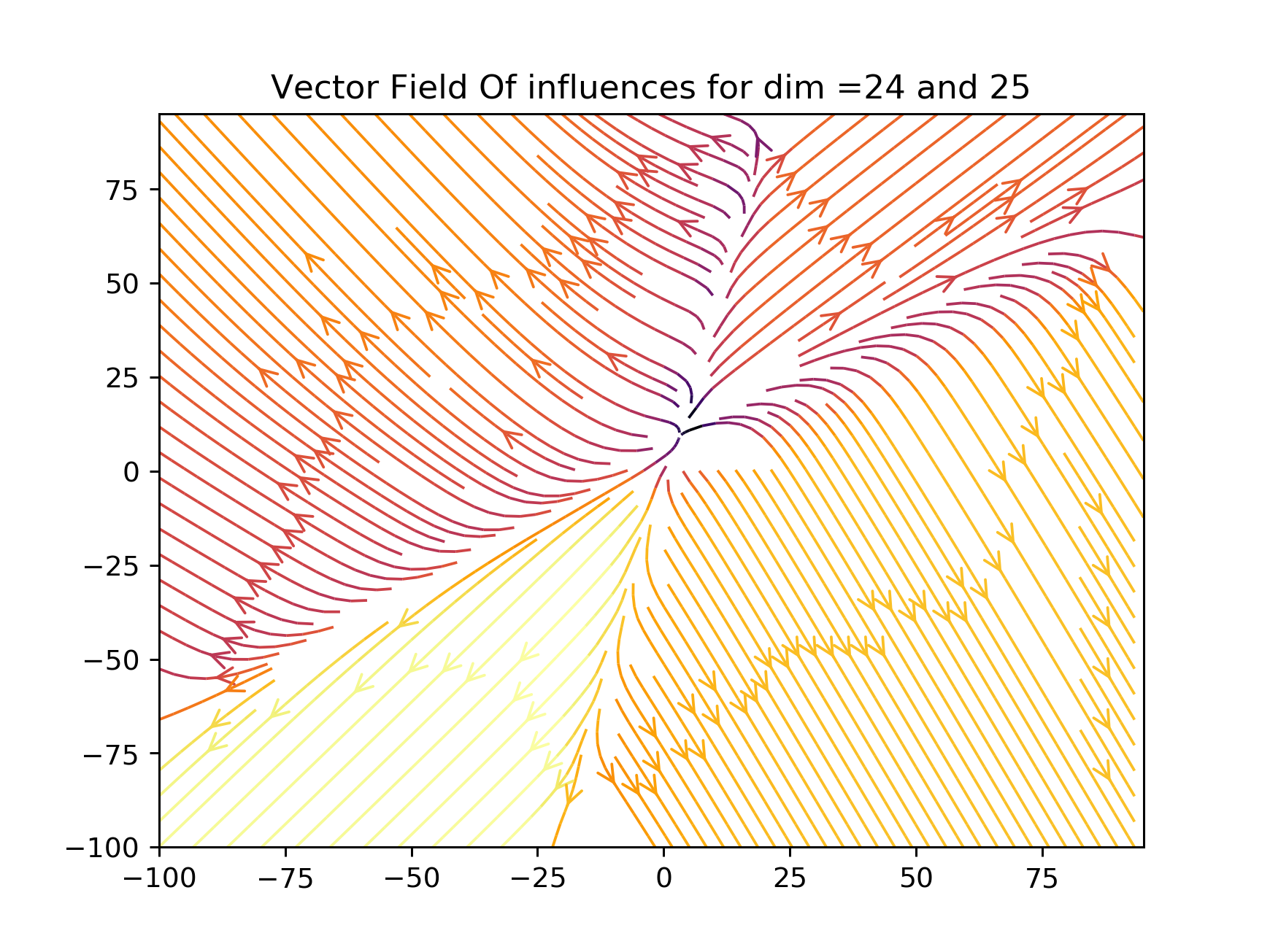}
		\caption{{\scriptsize few people influence many (source).}}
		\label{fig:12}
	\end{subfigure}%
	\begin{subfigure}[b]{0.31\textwidth}
		\includegraphics[width=\linewidth]{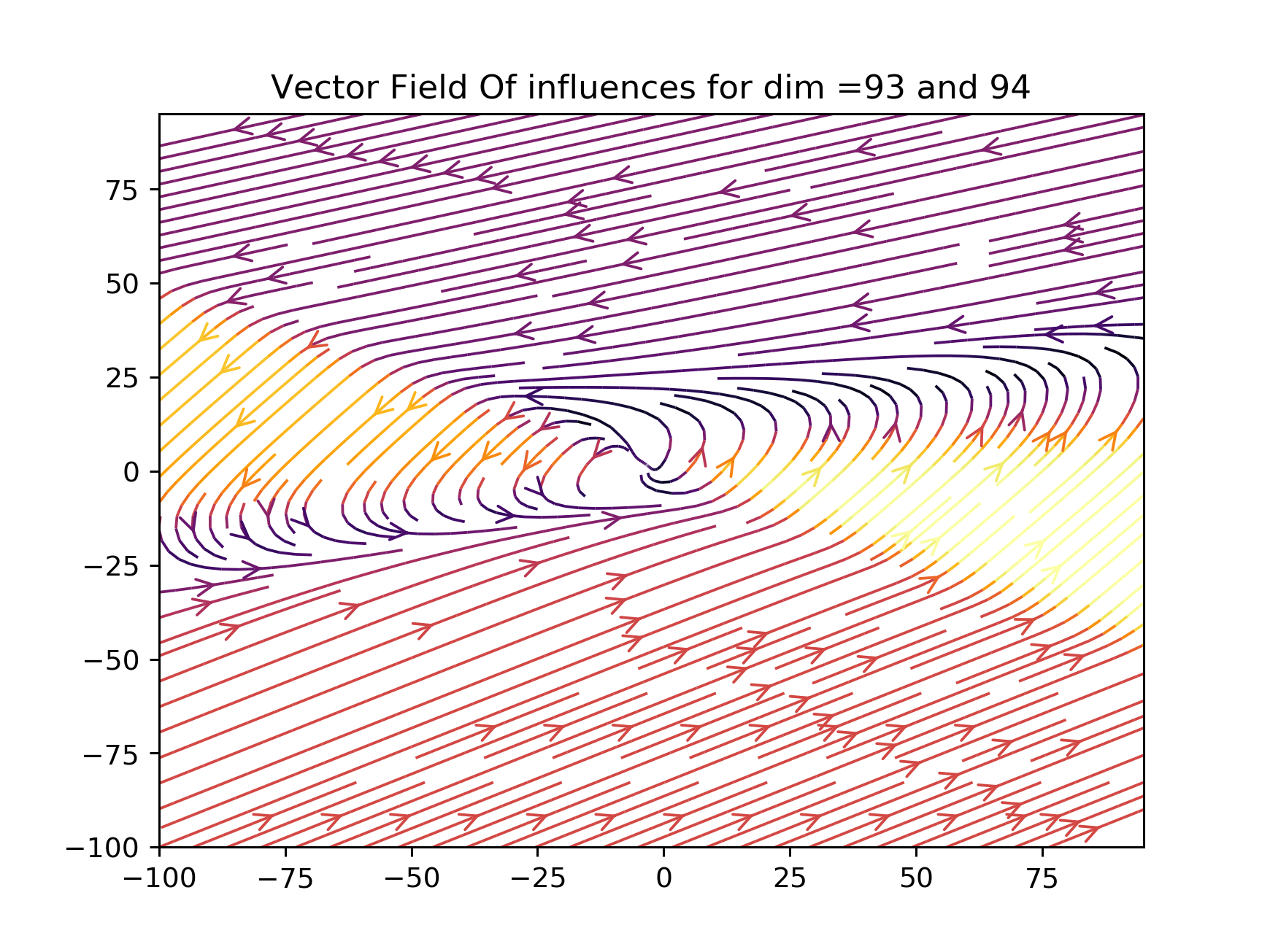}
		\caption{{\scriptsize dense influencing of many others (loop and path).}}
		\label{fig:13}
	\end{subfigure}%
	
	%\stepcounter{row}{}%
	\begin{subfigure}[b]{0.31\textwidth}
		\includegraphics[width=\linewidth]{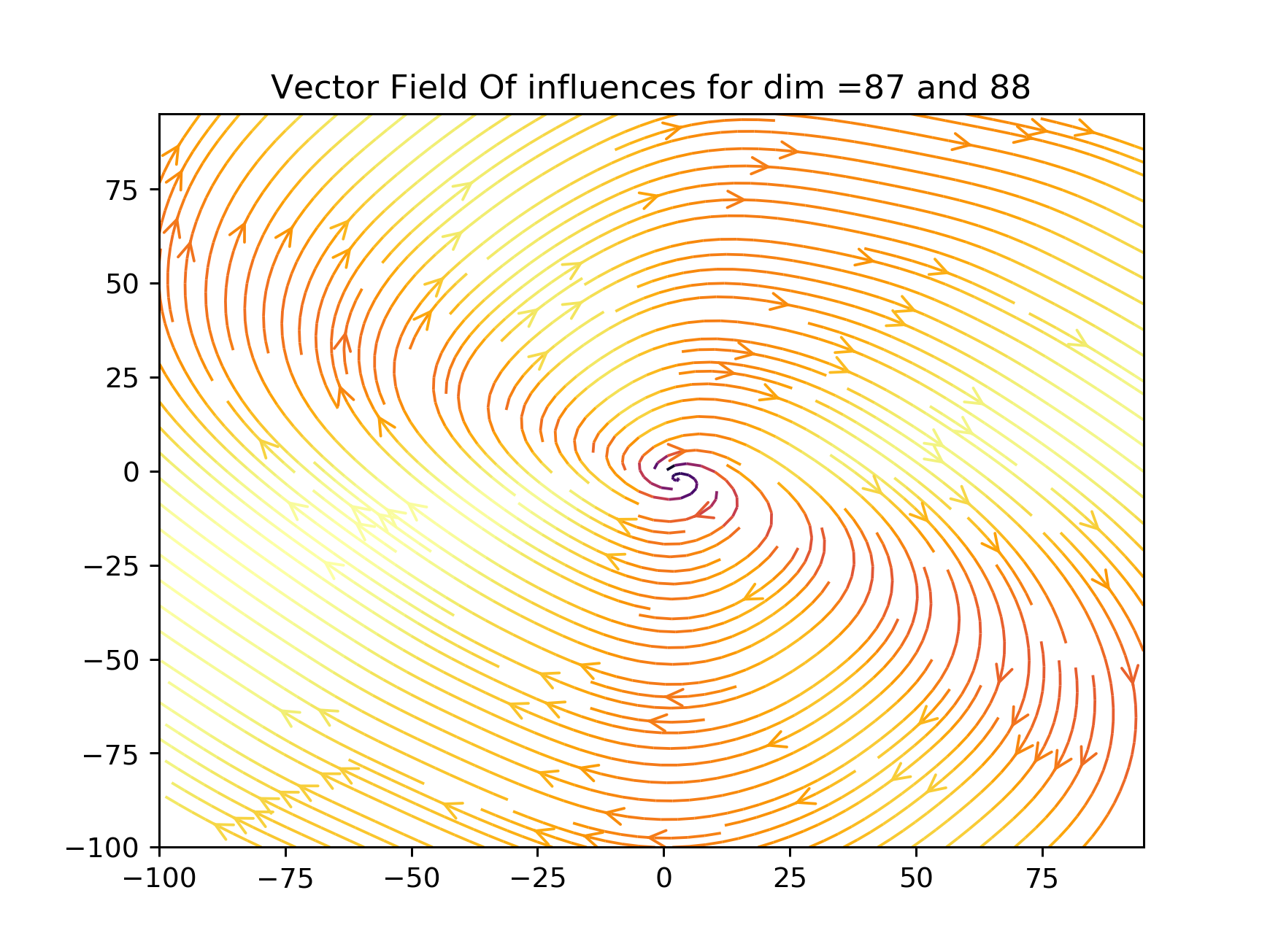}
		\caption{{\scriptsize many people influencing each other (loop)}}
		\label{fig:21}
	\end{subfigure}%  
	\begin{subfigure}[b]{0.31\textwidth}
		\includegraphics[width=\linewidth]{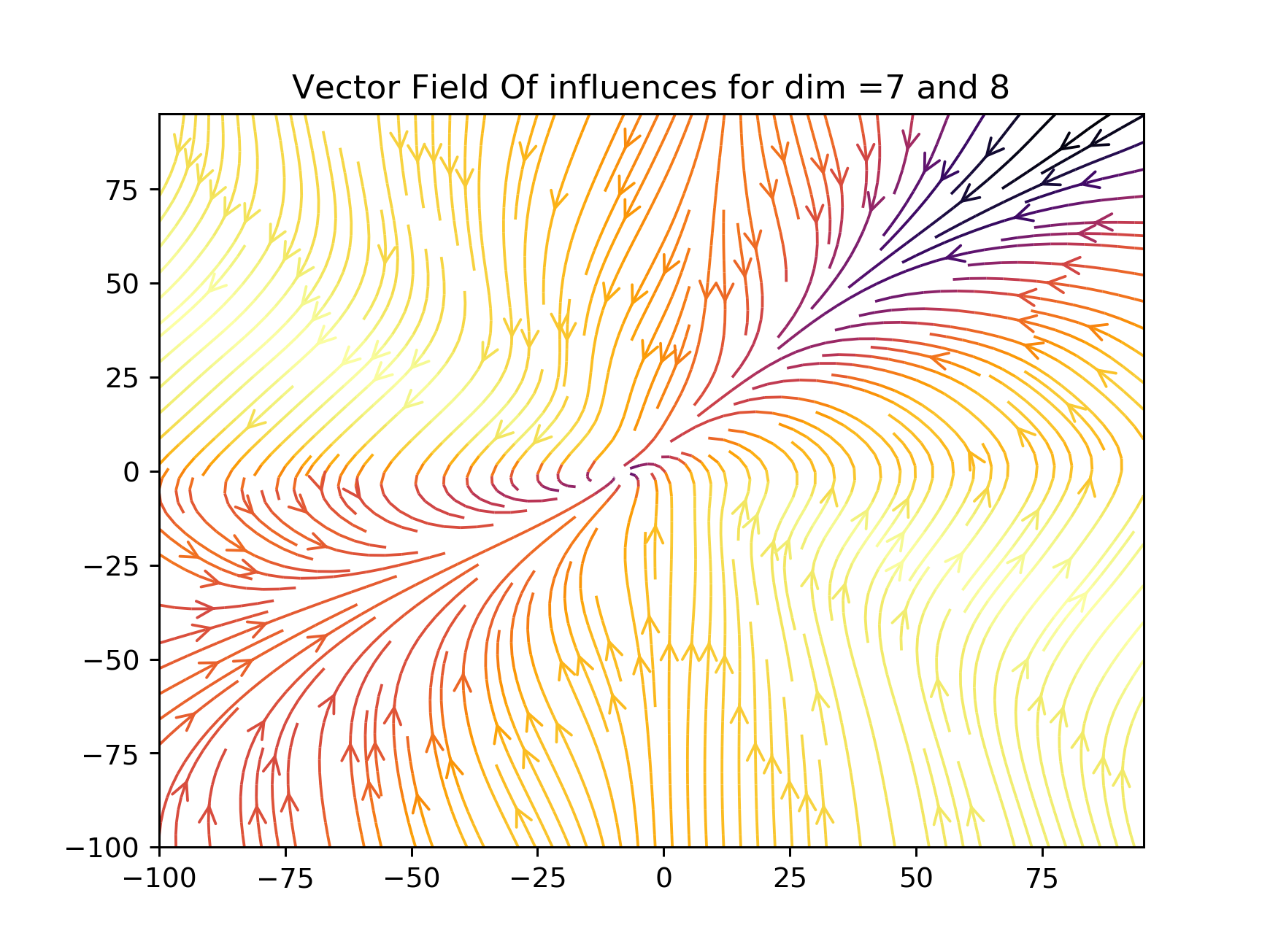}
		\caption{{\scriptsize few people being influenced by many (sink).}}
		\label{fig:22}
	\end{subfigure}%
	\begin{subfigure}[b]{0.31\textwidth}
		\includegraphics[width=\linewidth]{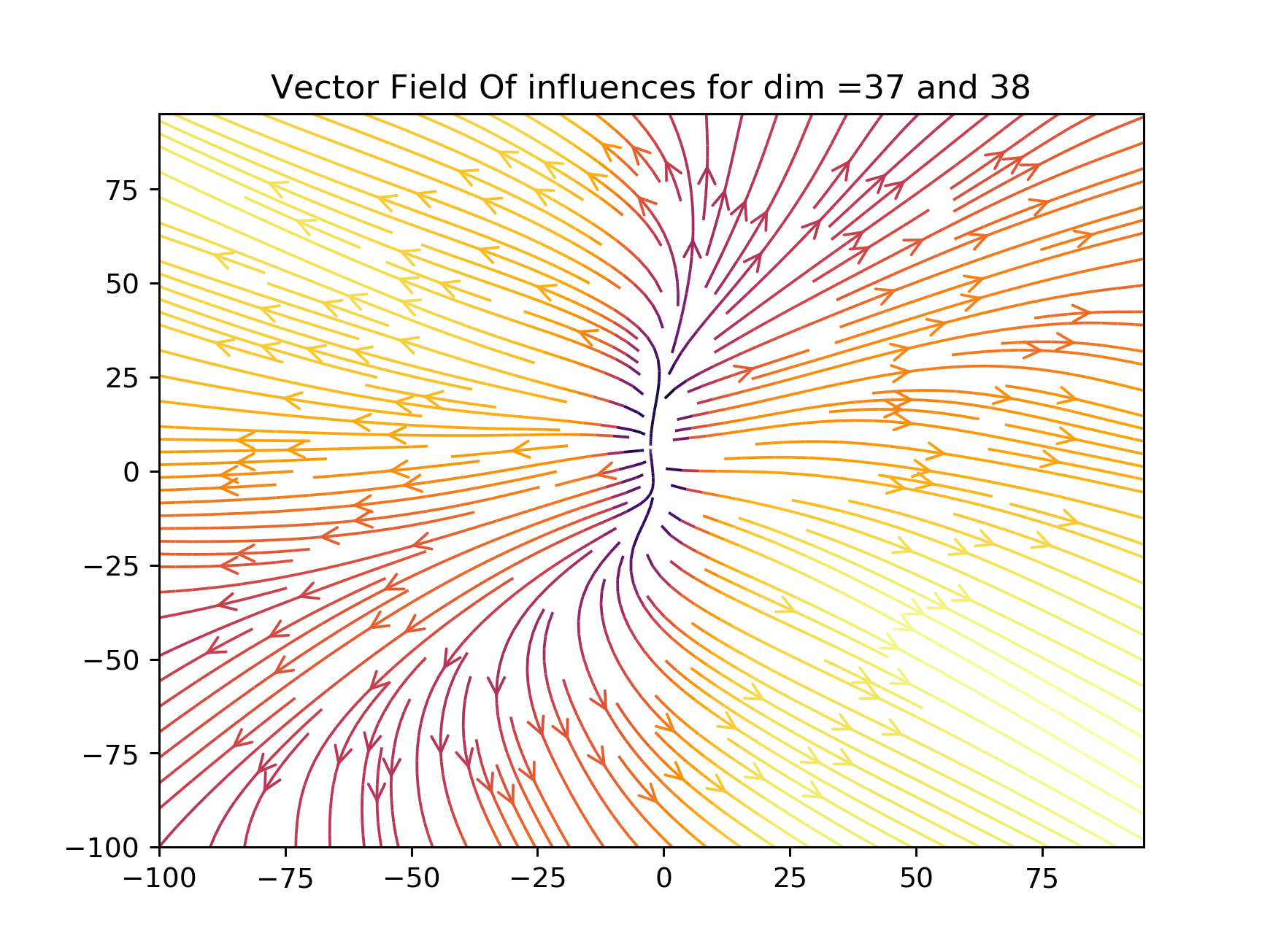}
		\caption{{\scriptsize density in few people influencing others (source).}}
		\label{fig:23}
	\end{subfigure}%
	\caption{Illustration of different vector fields for the relation of ``influences'' from the YAGO3-10 dataset. The X and Y axis correspond to the 2D dimensions of vector fields. }
	\label{fig:1}
\end{figure*}

\begin{figure*}[ht!]
	\centering 
	\begin{subfigure}[b]{0.31\textwidth}
		\includegraphics[width=\linewidth]{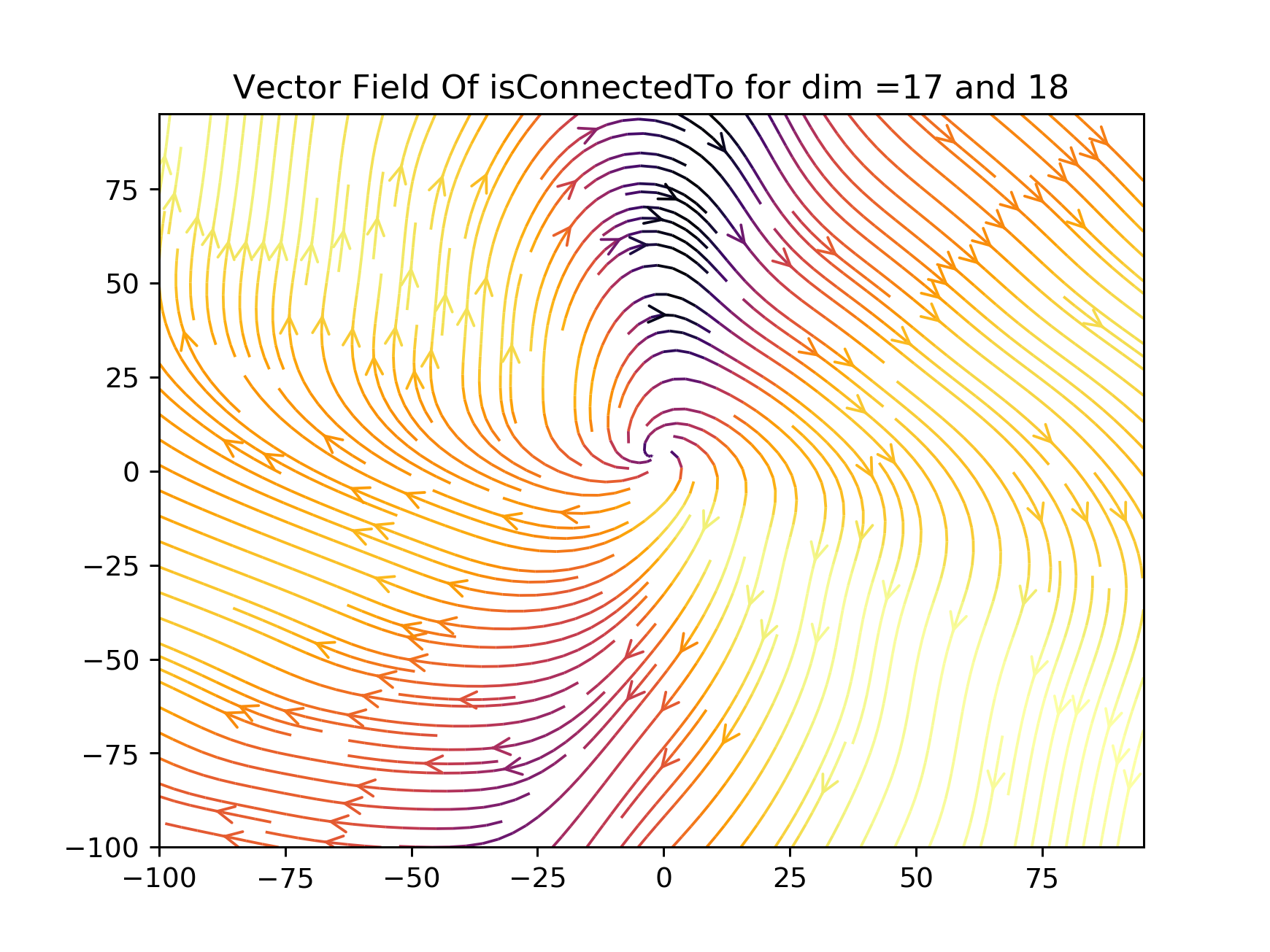}
	     \caption{{\scriptsize isconnectedto.}}
		\label{fig:isconnectedto1}
	\end{subfigure}%
	\begin{subfigure}[b]{0.31\textwidth}
		\includegraphics[width=\linewidth]{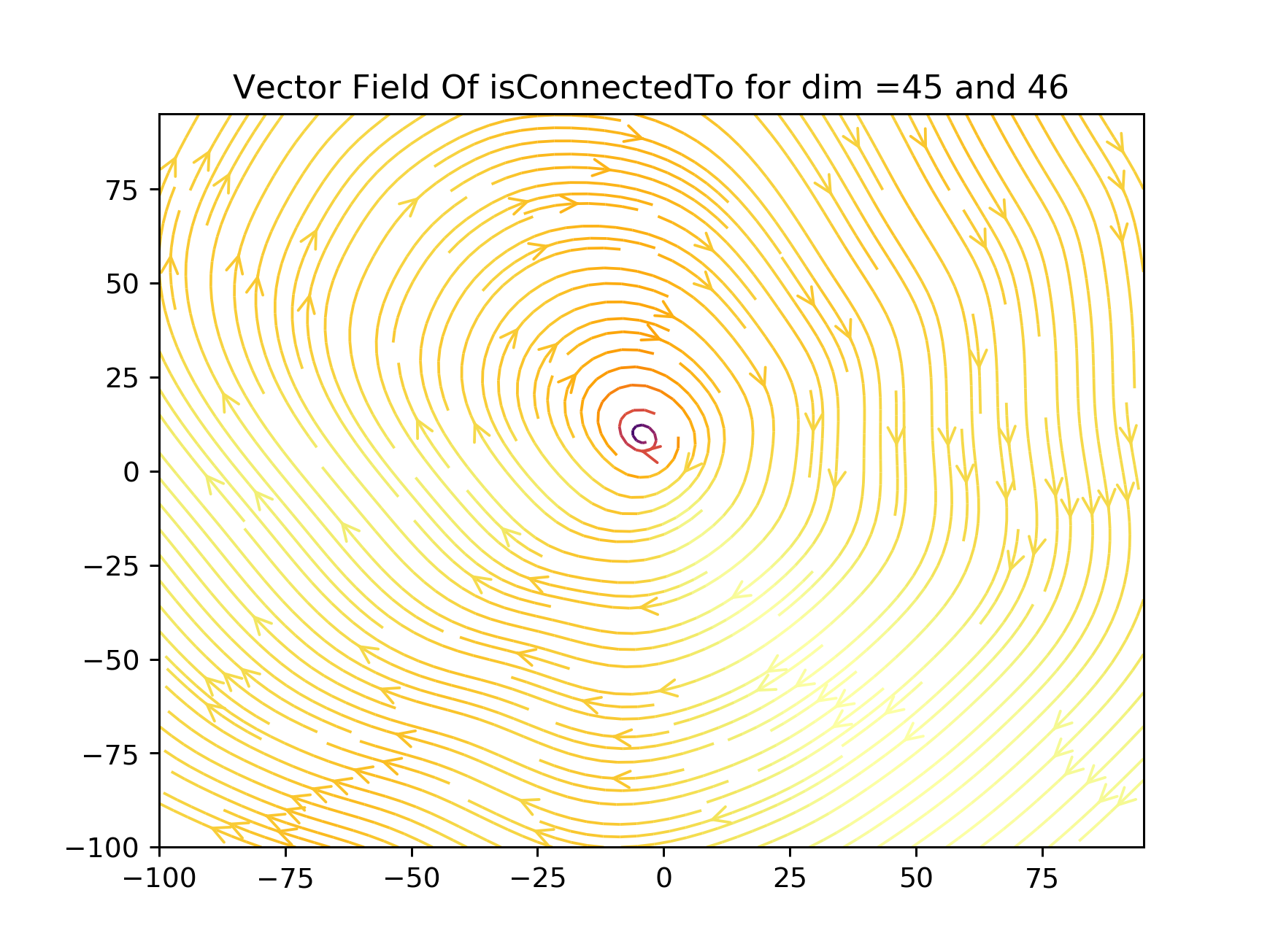}
		\caption{{\scriptsize isconnectedto.}}
		\label{fig:isconnectedto2}
	\end{subfigure}%
	\begin{subfigure}[b]{0.31\textwidth}
		\includegraphics[width=\linewidth]{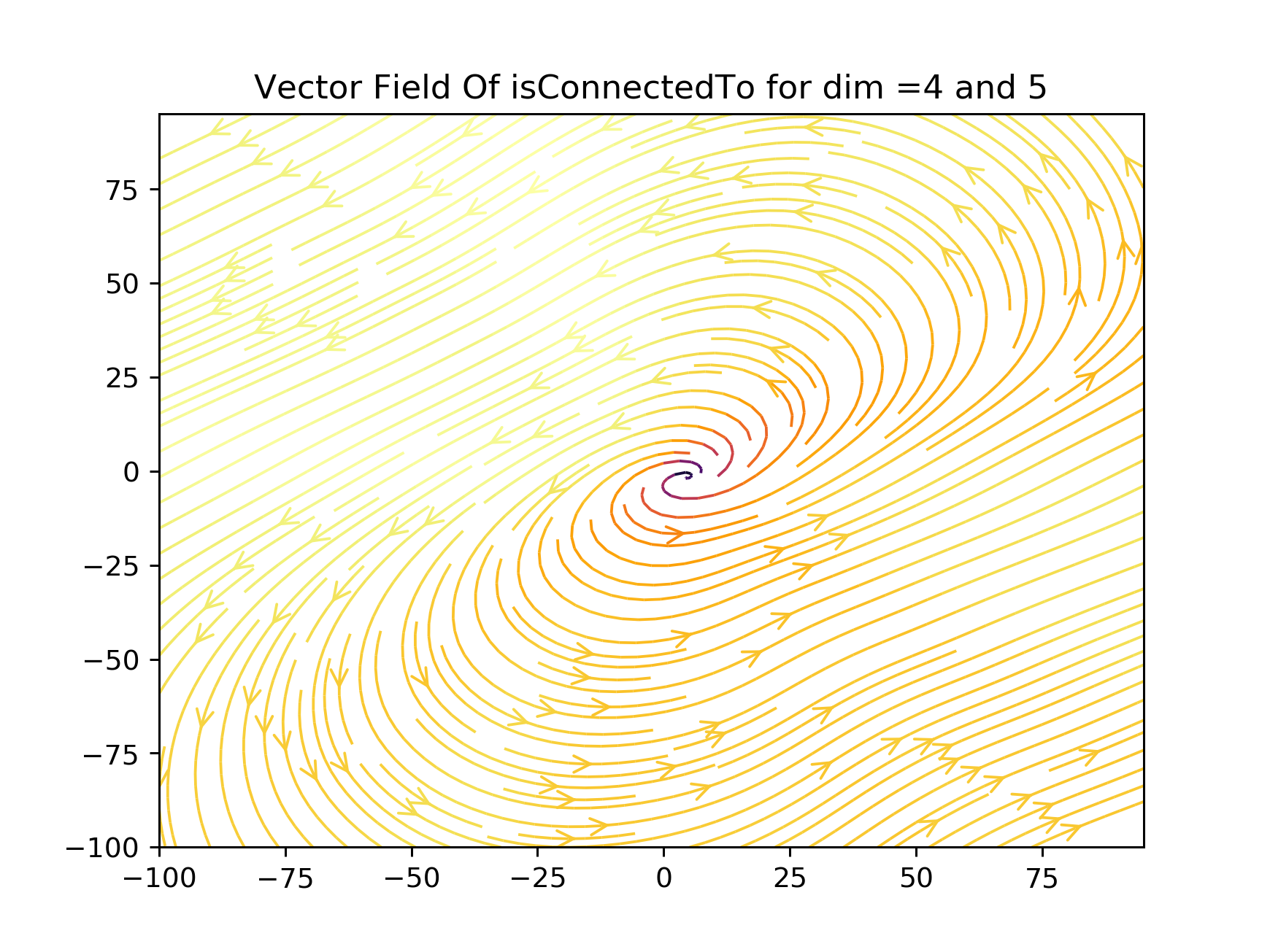}
		\caption{{\scriptsize isconnectedto.}}
		\label{fig:isconnectedto3}
	\end{subfigure}%
	
		%\stepcounter{row}{}%
	\begin{subfigure}[b]{0.31\textwidth}
		\includegraphics[width=\linewidth]{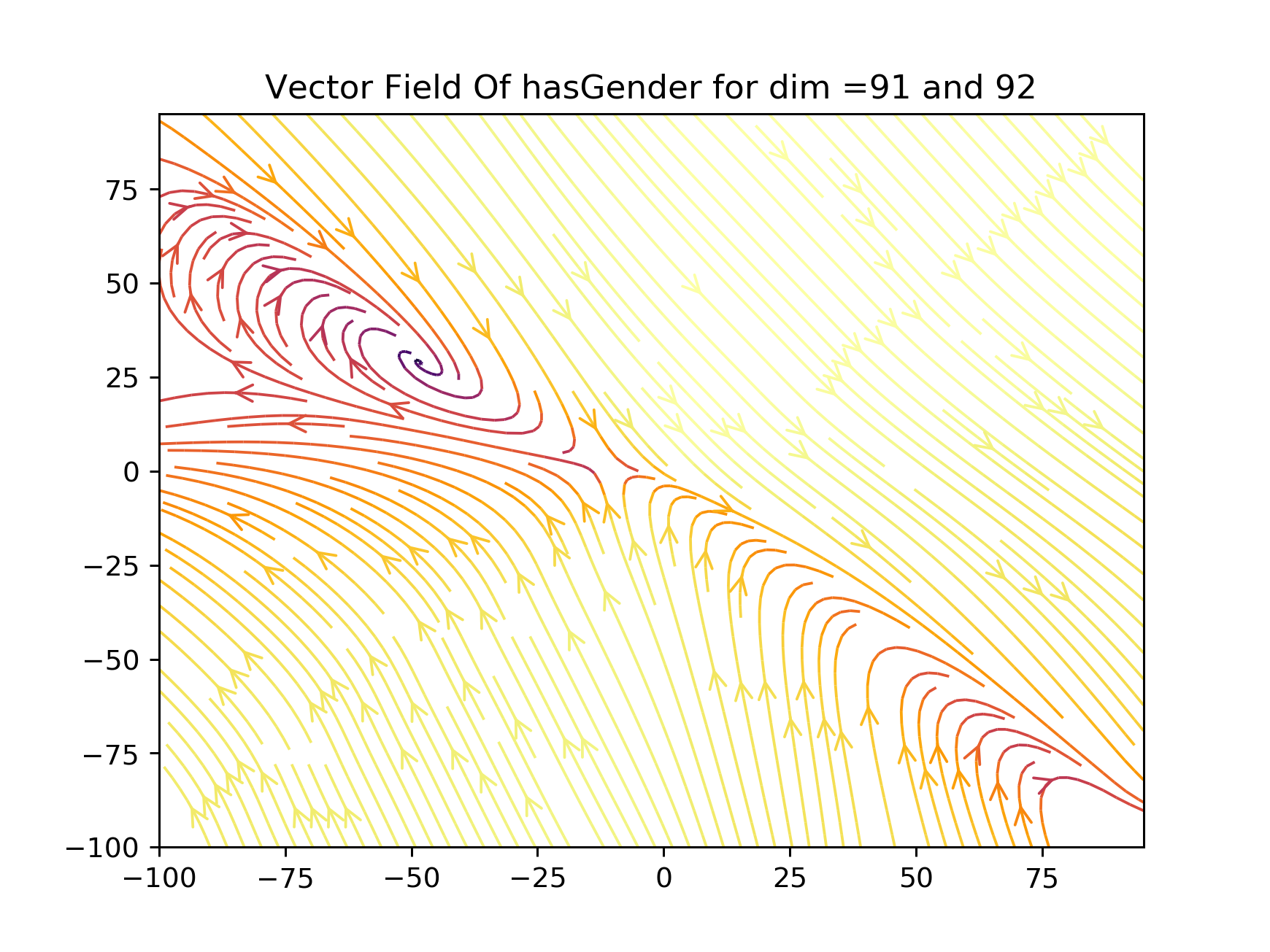}
		\caption{{\scriptsize hasGender}}
		\label{fig:hasGender1}
	\end{subfigure}%  
	\begin{subfigure}[b]{0.31\textwidth}
		\includegraphics[width=\linewidth]{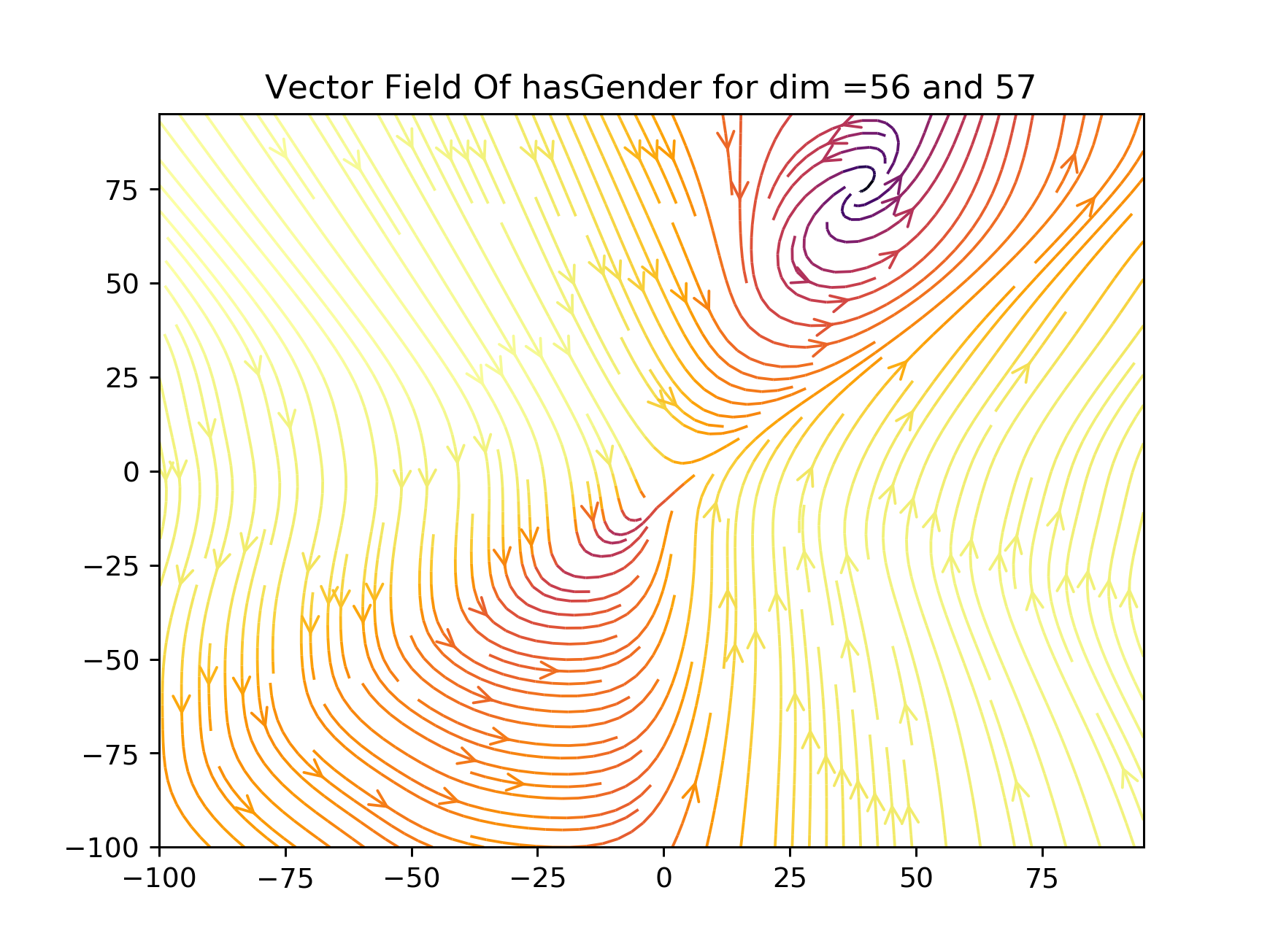}
		\caption{{\scriptsize hasGender.}}
		\label{fig:hasGender2}
	\end{subfigure}%
	\begin{subfigure}[b]{0.31\textwidth}
		\includegraphics[width=\linewidth]{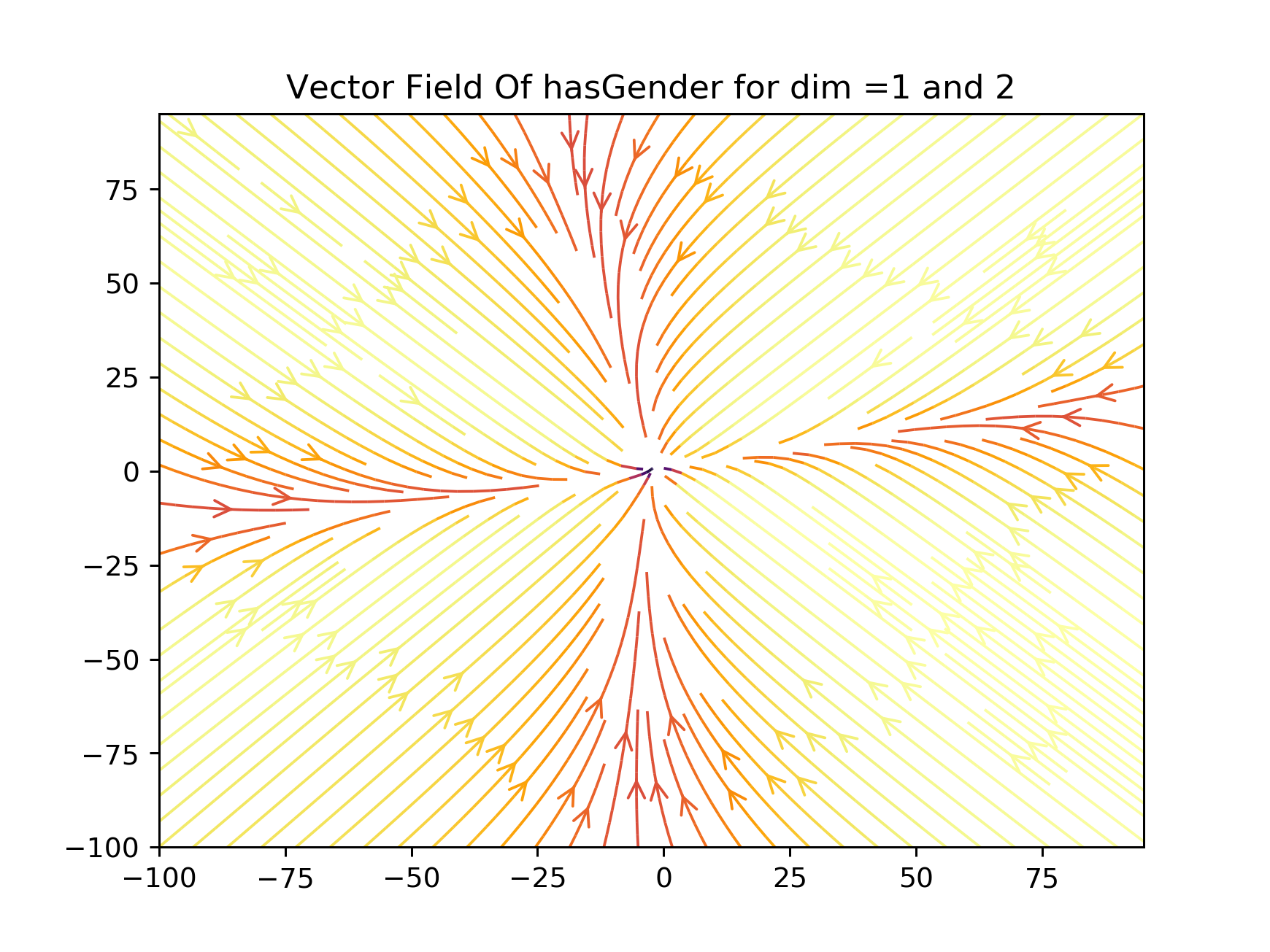}
		\caption{{\scriptsize hasGender.}}
		\label{fig:hasGender3}
	\end{subfigure}%
	
			%\stepcounter{row}{}%
	\begin{subfigure}[b]{0.31\textwidth}
		\includegraphics[width=\linewidth]{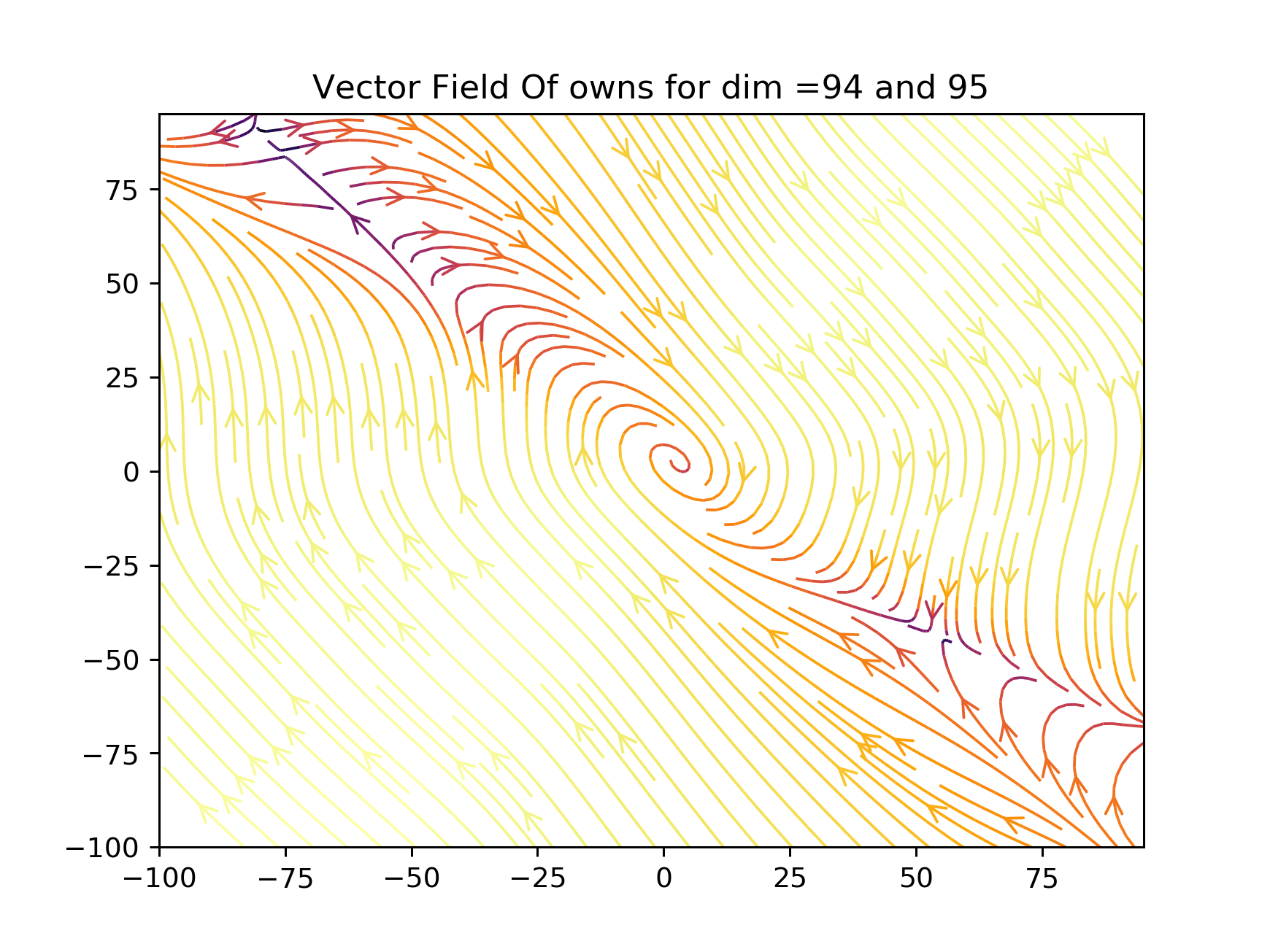}
		\caption{{\scriptsize owns}}
		\label{fig:owns1}
	\end{subfigure}%  
	\begin{subfigure}[b]{0.31\textwidth}
		\includegraphics[width=\linewidth]{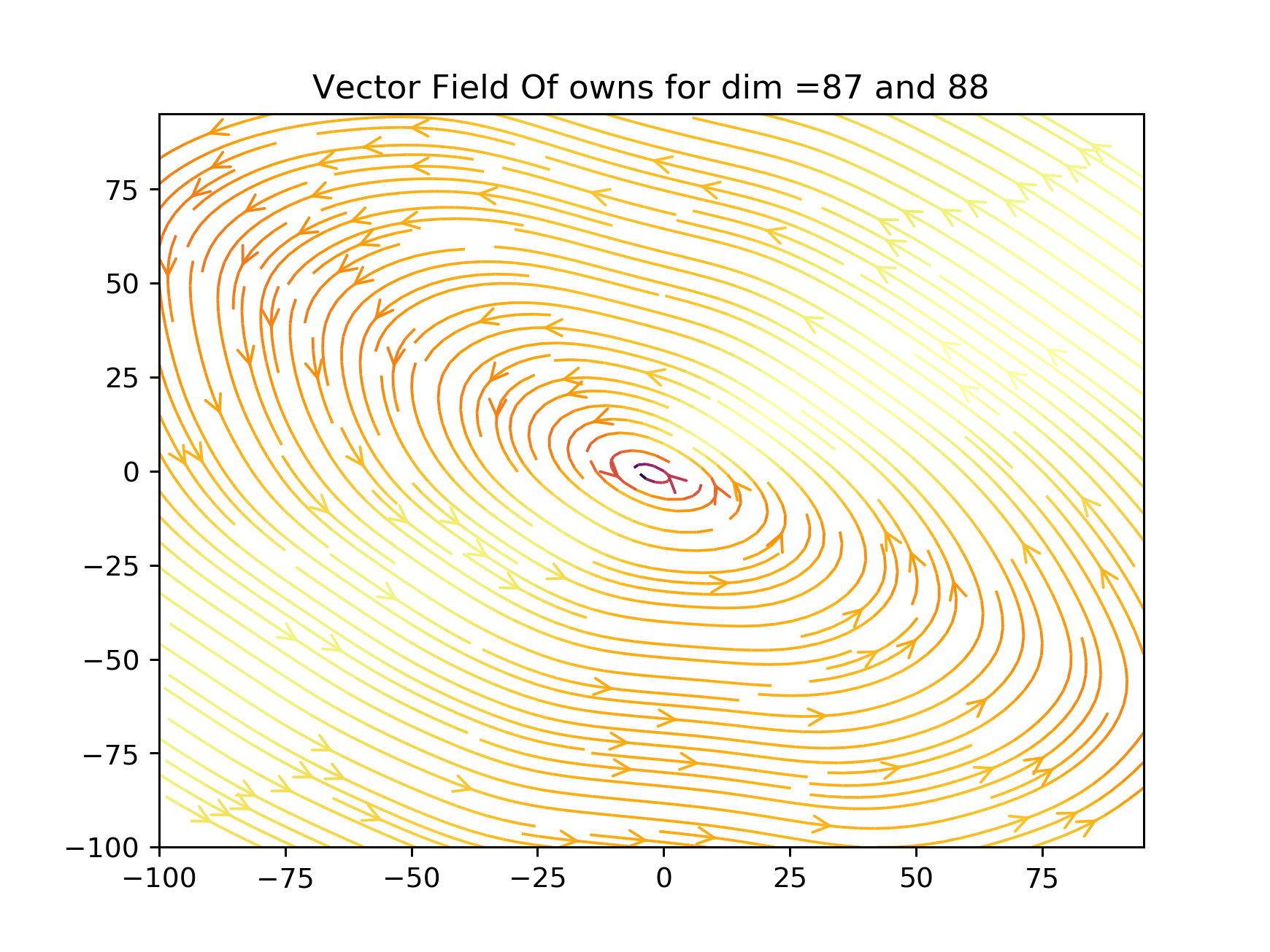}
		\caption{{\scriptsize owns.}}
		\label{fig:owns2}
	\end{subfigure}%
	\begin{subfigure}[b]{0.31\textwidth}
		\includegraphics[width=\linewidth]{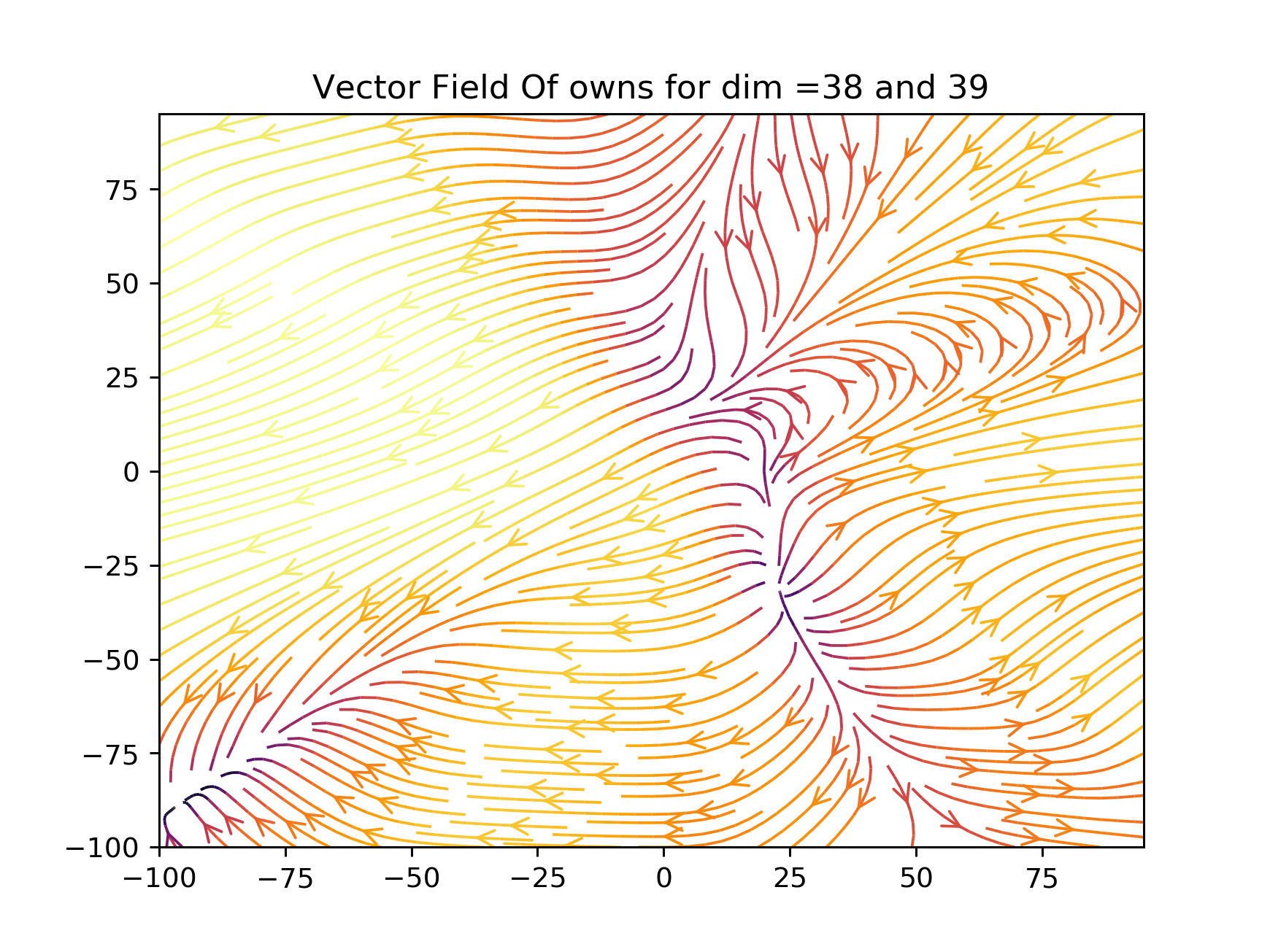}
		\caption{{\scriptsize owns.}}
		\label{fig:owns3}
	\end{subfigure}%
	
			%\stepcounter{row}{}%
	\begin{subfigure}[b]{0.31\textwidth}
		\includegraphics[width=\linewidth]{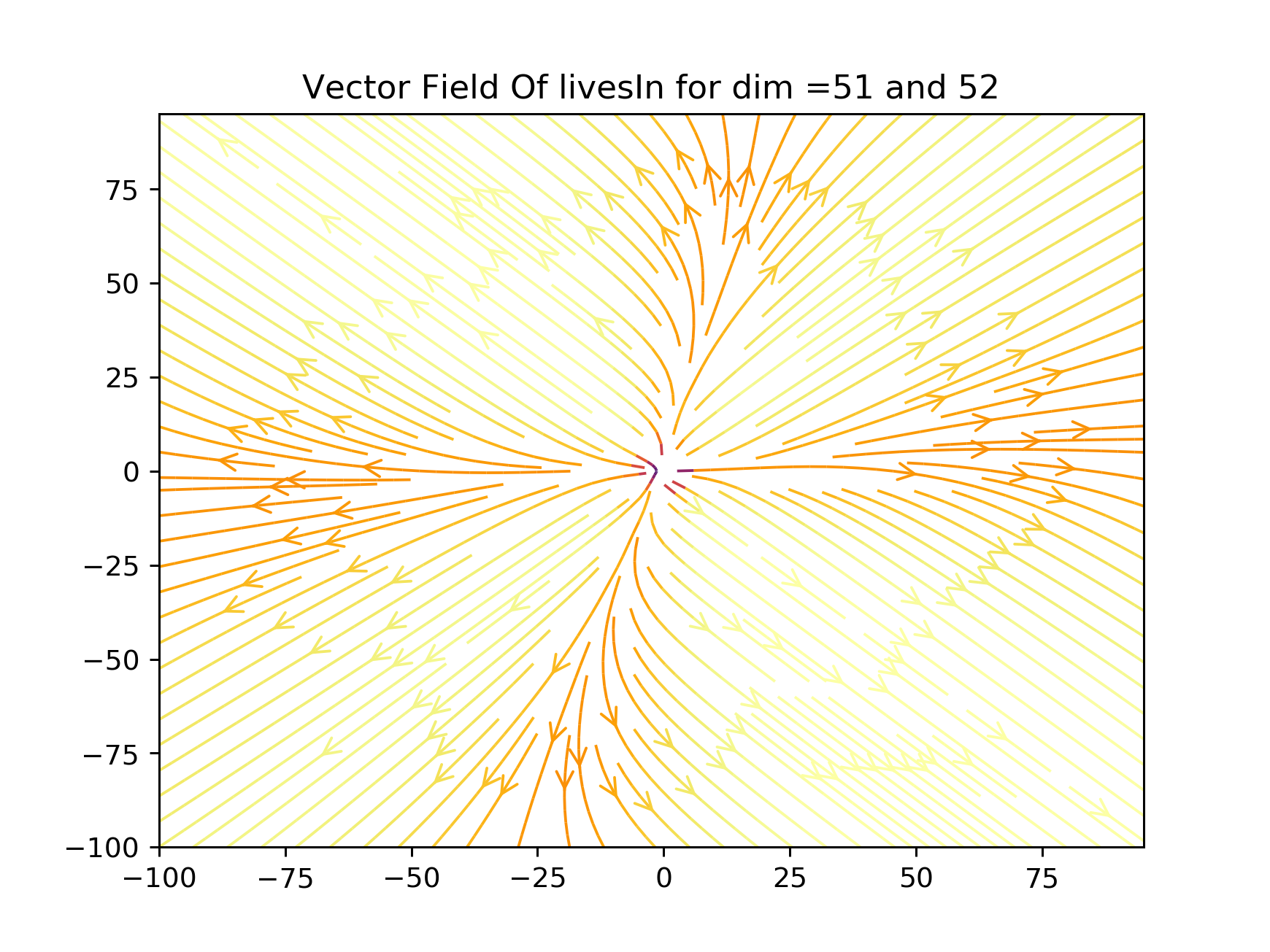}
		\caption{{\scriptsize livesIn}}
		\label{fig:livesIn1}
	\end{subfigure}%  
	\begin{subfigure}[b]{0.31\textwidth}
		\includegraphics[width=\linewidth]{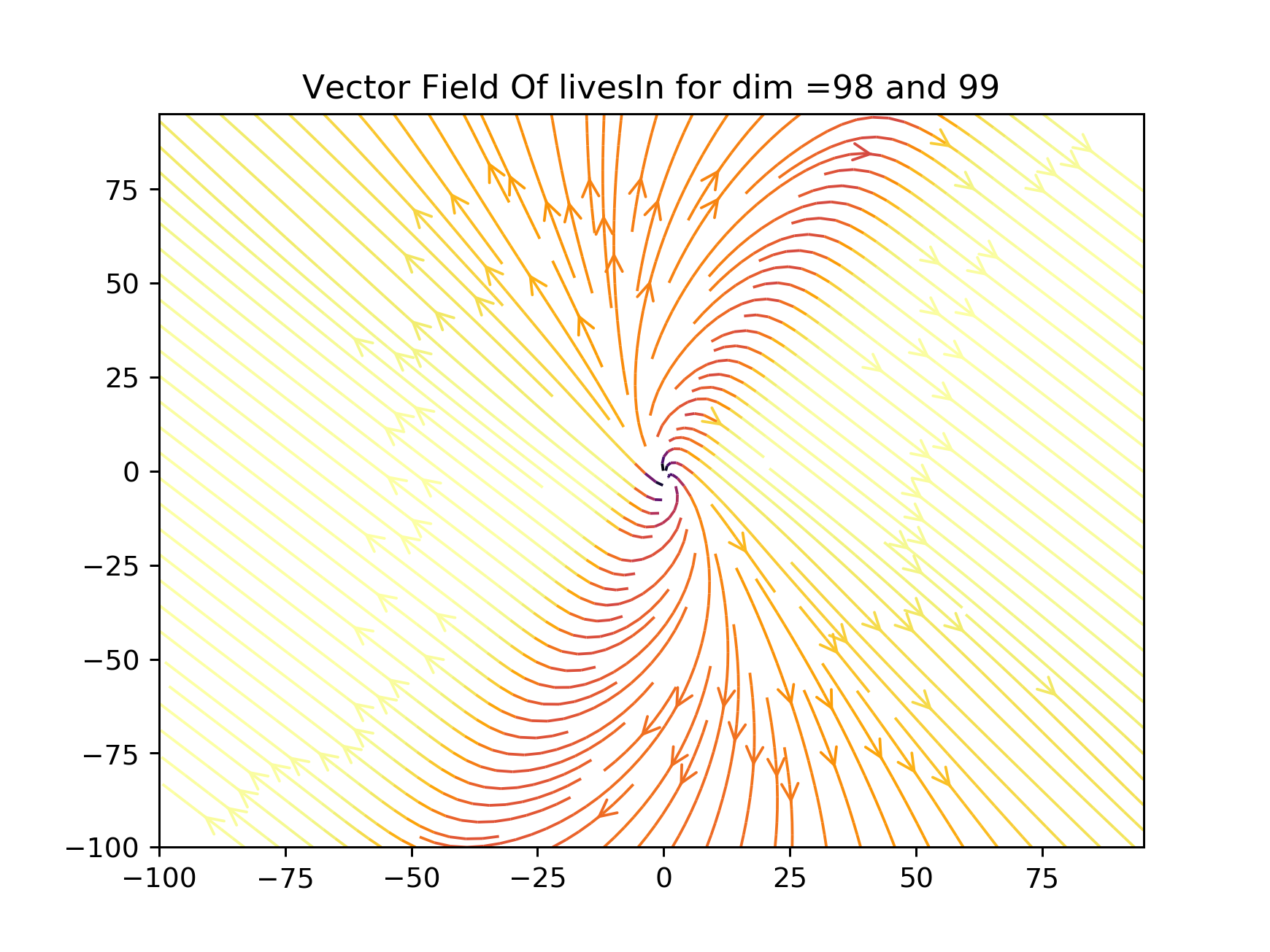}
		\caption{{\scriptsize livesIn.}}
		\label{fig:livesIn2}
	\end{subfigure}%
	\begin{subfigure}[b]{0.31\textwidth}
		\includegraphics[width=\linewidth]{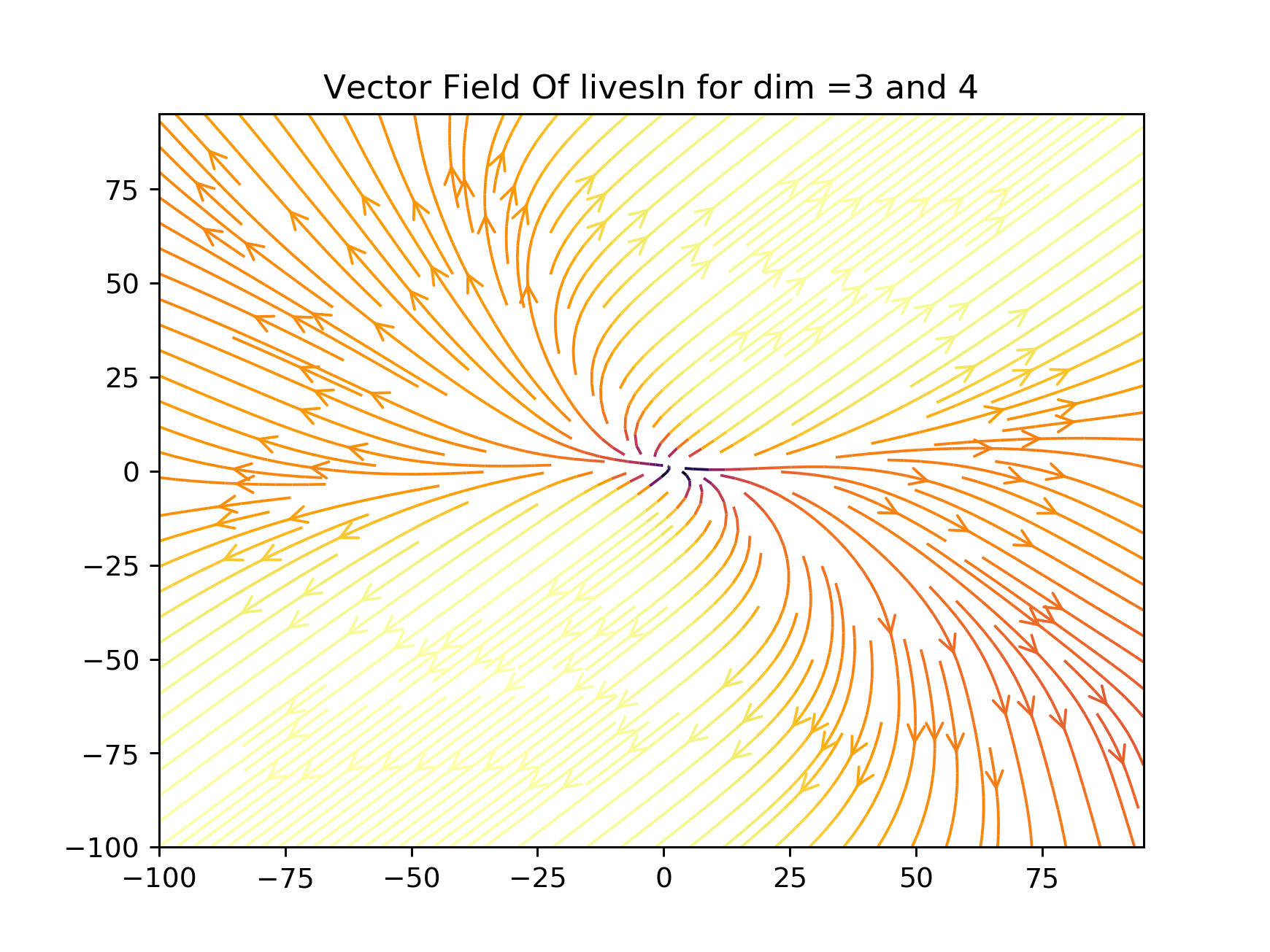}
		\caption{{\scriptsize livesIn.}}
		\label{fig:livesIn3}
	\end{subfigure}%
	\caption{Illustration of vector fields learned for different relations.}
	\label{fig:2}
\end{figure*}

The results are shown in Table \ref{table:result_table_wn} and the illustrations are depicted in Figure \ref{fig:1}, and Figure \ref{fig:2}.
We first report the performance comparison of \emph{FieldE} and other models. 
% The results of RotatE and ComplEx as well as the results of TransE in FB15k-237 are taken from the original RotatE paper \cite{sun2019rotate}. We took the results of TuckEr, QuatE, Dismult, ConvE, and MuRP from their original papers. The results for TransE on YAGO3-10, are computed by our reimplementation.
On both of the datasets, \emph{FieldE} outperforms all the other models on all the metrics. 
On the FB15k-237 dataset, except MRR achieved equally by the Tucker model, all the other models fall short in comparison to \emph{FieldE}. 
FieldE also outperform all the models in MRR on YAGO3-10.

% \begin{figure*}%
%     \centering
%     \subfloat[test]{{\includegraphics[trim=0cm 0 0cm 0, clip,width=6cm,height=5.5cm]{motifs/4142-YAGO3-10.png} }}%
%     \qquad
%     \subfloat[test]{{\includegraphics[trim=0cm 0 0cm 0, clip,width=6cm,height=5.5cm]{motifs/4142-YAGO3-10.png}}}%
%     \qquad
%     \subfloat[test]{{\includegraphics[trim=0cm 0 0cm 0, clip,width=6cm,height=5.5cm]{motifs/4142-YAGO3-10.png}}}%
%     \caption{test.}%
% \label{fig: Semantic RodE}
% \end{figure*}

\begin{table*}[ht!]
\caption{Link prediction results on d FB15k-237, and YAGO3-10. The highlight of performances for different models are marked. }
\begin{tabular}{lllllllll}
 \toprule 
   \multirow{1}{*}{Model} & \multicolumn{4}{c}{FB15k-237}         & \multicolumn{4}{c}{YAGO3-10}      \\ \cline{1-1} \cline{2-9} 
                          & MRR & Hits@1 & Hits@3 & Hits@10  & MRR & Hits@1 & Hits@3 & Hits@10 \\ \cline{2-9} 
                 TransE   & 0.33 &   0.23   &   0.37  &  0.53 & 0.49 & 0.39 & 0.56 & 0.67 \\ \cline{2-9} 
                 RotatE      & 0.34 &   0.24   &    0.37  &  0.53 &  0.49 &  0.40 & 0.55 &  0.67 \\   \cline{2-9}
                 %.949 .944 .952 .959
                TuckEr       & \cellcolor{blue!12}0.36 &   0.26  &   0.39  &  0.54 & -  & - & - &  -  \\ \cline{2-9} 
                ComplEx      & 0.32 &   -  &   -  &  0.51  & 0.36  & 0.26 & 0.40 & 0.55    \\ \cline{2-9} 
                QuatE        & 0.31 &   0.23  &   0.34  &  0.49 & -  & - & - &  - \\ \cline{2-9} 
               % SimplE    &-    & -     & -     & -    &-    & -     & -     &-       \\ \cline{2-9} 
                Dismult      & 0.24  & 0.15   & 0.26   & 0.42 & 0.34  & 0.24 & 0.28 &   0.54  \\ \cline{2-9}
                ConvE        & 0.34  &  0.24  &  0.37  &  0.50 & 0.44 & 0.35 &  0.49 &  0.62   \\ \cline{2-9}
                MuRP         & 0.34  &  0.25  & 0.37 &  0.52 & 0.35  & 0.25 & 0.40 &  0.57 \\  
\hline
     \bottomrule 
                FieldE      &  \cellcolor{blue!12}0.36 &\cellcolor{blue!12}0.27  &  \cellcolor{blue!12}0.39 & \cellcolor{blue!12}0.55 & \cellcolor{blue!12}0.51 & \cellcolor{blue!12}0.41 &\cellcolor{blue!12} 0.58 & \cellcolor{blue!12}0.68 \\ 
\hline
\end{tabular}
%\end{adjustbox}
\label{table:result_table_wn}
\end{table*}

Following our motivating example which was focused with motifs created by the \emph{influences} relation in the YAGO3-10 knowledge graph, we provide sample visualizations for vector fields of this relation in Figure \ref{fig:1}. 
These visualizations have been selected among hundreds to only give an impression about the capability of FieldE model in motif learning. 
In order to provide a presentable illustration, we plot each vector fields in pair of dimensions. Therefore, for FieldE with $d=100$, we created $99$ pairs among which we selected six graphs constructed from dimension $\{(\boldsymbol{e}_7,\boldsymbol{e}_8), (\boldsymbol{e}_{24},\boldsymbol{e}_{25}), (\boldsymbol{e}_{37},\boldsymbol{e}_{38}),(\boldsymbol{e}_{41},\boldsymbol{e}_{42}), (\boldsymbol{e}_{87},\boldsymbol{e}_{88}), (\boldsymbol{e}_{93},\boldsymbol{e}_{94})\}$.

In subfigure \ref{fig:11}, the captured vector fields are shaping both as loop and path motifs simultaneously. 
It explains exactly the case illustrated in Figure \ref{fig:path} where some people are influencing others in a loop structure, and some people influence others in a path structure (without a return link). 
This shows a full structure preservation from the graph representation to the vector representation. 
As discussed before, this capability also avoids wrong inferences. 
Subgraph \ref{fig:12} shows trajectories of some people being a \emph{source} influencer for many others. 
The subfigure \ref{fig:13} is another loop and path occurrence with more density. 

A set of vector fields with a lot of loops is illustrated in subfigure \ref{fig:21}. 
The interpretation of this vector field is that, there are a series of different people influencing each other in a loop structure with different number of entities. 
The subfigure \ref{fig:22} shows a set of sink nodes where they have been influenced by many.
And finally, the subfigure \ref{fig:23} shows some more dense source entities. 
Overall, these illustrations double-prove the capability of \emph{FieldE} inherited from ODEs and facilitated by the concept of vector field and trajectories. 

The visualizations in Figure \ref{fig:2} represent the subgraphs with different motifs including path and loop in different relations. 
Each row corresponds to the illustrations of one relation for which three different learned structures are selected to be shown.
For example, subgraphs of \ref{fig:isconnectedto1}, \ref{fig:isconnectedto2}, and \ref{fig:isconnectedto3} correspond to the ``isconnectedto'' relation that shows which airports are connected to each other in different structures (loop and path). 
Our visualizations capture different learned motifs including path, and loop which show some airports are connected in a loop form and some not. 
%%Second set of visulizations (sungraphs of \ref{fig:hasOfficialLanguage1}, \ref{fig:hasOfficialLanguage2}, and \ref{fig:hasOfficialLanguage3}) belong to ``hasOfficialLanguage'' relation. 
%These vector fields show people with one or several official languages. 
In the subgraphs of \ref{fig:hasGender1}, \ref{fig:hasGender2}, \ref{fig:hasGender3}, different motifs of ``hasGender'' relation are captured. 
Same for the ``livesIn'' relation, we show different illustrations of the vector fields in \ref{fig:livesIn1}, \ref{fig:livesIn2}, and \ref{fig:livesIn3}.
By all of these illustration, we aim at giving clarity on the effect of ODEs in learning vector fields which avoids wrong inferences. 
All of these are trajectories lain on relation-specific Reimanian manifold learned by the neural network of our model. 
The arrows shows the direction of motif evolution in the vector space for each shape.

%% file: conclusion.tex
\section{Conclusion}
This work presented a novel embedding models \emph{FieldE} which is designed based on Ordinary Differential Equations.
Since it inherits the characteristics of ODEs, it is capable of encoding different semantical and structural complexities in knowledge graphs.
We modeled relations as vector fields on a Rimannian manifold and the entities of a knowledge graph which are connected through the considered relation, are taken as points on the trajectories laid on the manifold. We specifically parameterize the vector field by a neural network to learn the underlying geometry from the training graph.
We examined \emph{FieldE} on several datasets and compared it with a selection of best state-of-the-art embedding models. 
\emph{FieldE} majorly outperforms all the models in all the metrics. 
%It lives up the premise of ODEs in preserving the complex structure of subgraphs. 
We focused on showing the motif learning for loop and path simultaneously and preserving their structures from the graph representation to the vector representation. 
%None of the already designed embedding models have been seen the use of trajectories as relations and the points of vector fields as entities of a graph. 
%This research have been performed on the link prediction task and can further extended to classification.
We showed that the neural network of \emph{FieldE} learns various shapes of the vector fields and consequently the underlying geometry.  
In future versions of this, we plan to apply it on real world knowledge graph beside YAGO and FreeBase and further explore the effect of ODEs in learning process of knowledge graph embedding models.